%% file: main.tex
\documentclass{article} % For LaTeX2e
\usepackage{iclr2027_conference,times}

\input{math_commands.tex}

\usepackage{booktabs}
\usepackage{graphicx}
\usepackage{hyperref}
\usepackage{makecell}
\usepackage{url}

\title{An Empirical Study of Reinforcement Learning Scaling Laws}

\author{Zeyu Zheng, Clare Lyle, Khimya Khetarpal \& Will Dabney \\
Google DeepMind \\
\texttt{\{zeyuzheng,clarelyle,khimya,wdabney\}@google.com}
}

\newcommand{\cutcaptiondown}{\vspace{-0.0in}}

\iclrarxiv % Uncomment for preprint version, but NOT for submission.
\begin{document}

\maketitle

\begin{abstract}
The discovery of scaling laws enables prediction of \textit{the performance}, \textit{the optimal resource allocation strategy}, and \textit{the best hyperparameter configuration} of a large-scale training job based on inexpensive experiments at smaller scales. 
Guided by these empirical scaling laws, the deep learning community has achieved great success in training large (self-)supervised models on Internet-scale datasets. 
In contrast, despite its potential and wide application, reinforcement learning (RL) has yet to find a guiding principle for predictable scaling. 
In this work, we conduct an empirical study of RL scaling laws of an actor-critic algorithm in two procedurally generated environments, Starpilot and Craftax. 
We discover empirical scaling laws with strong predictive performance in both environments. The discovered scaling laws accurately predict the aforementioned three quantities for larger-scale training jobs up to two orders of magnitude in total training compute. 
In particular, we employ a variant of Elo scores for the single-agent RL paradigm as an alternative performance metric than the commonly used mean episode return. Our results show that the Elo score is a much more predictable metric than the mean episode return, making it a better choice for fitting scaling laws.
By applying our discovered scaling laws to increase training compute, we establish a new state of the art in the challenging Craftax environment. 
Beneath the surface of the global scaling trend, we also observe a non-monotonic scaling pattern of the compute-optimal model size in Craftax. We connect this observation to a pathology known as Ray interference.
\end{abstract}

\input{introduction}
\input{related_work}
\input{method}
\input{experiment}
\input{conclusion}

\subsection*{AI use statement}
In this work, we used generative AI tools for \textbf{implement methods, design or provide feedback on research  methodology or experiments, clean and reformat dataset, interpret results}.
We have not used generative AI tools for generate synthetic data sets, help develop theoretical models or conceptual framework, propose or refine hypotheses
and formulate mathematical claims, provide critical ingredients for proving mathematical claims, assist in the writing of proofs, assist with translation, support qualitative and thematic data analysis. are not applicable to this work.
Additionally, we used generative AI tools for \textbf{edit a research paper to improve readability}. We have reviewed all AI-assisted work. We reviewed the AI-generated code for experiment and data analysis. We proofread the text revised by AI tools. 
We take responsibility for the final content of this work,
including text, claims or artifacts produced with the aid of generative AI.

\ifshowacks
\subsubsection*{Acknowledgments}
We thank Dan A. Calian and Satinder Singh for their valuable feedback on an early version of this manuscript.
\fi

\bibliography{iclr2027_conference}
\bibliographystyle{iclr2027_conference}

\clearpage
\appendix
\input{appendix}

\end{document}

%% file: math_commands.tex
\usepackage{amsmath,amsfonts,bm}

\def\eqref#1{equation~\ref{#1}}
\def\1{\bm{1}}

\DeclareMathAlphabet{\mathsfit}{\encodingdefault}{\sfdefault}{m}{sl}
\SetMathAlphabet{\mathsfit}{bold}{\encodingdefault}{\sfdefault}{bx}{n}

%% file: introduction.tex
\section{Introduction}
\label{sec:intro}

The discovery of empirical scaling laws has revolutionized deep learning, in particular for generative modeling, by transforming model development from a process of heuristic trial-and-error into a \textit{predictable} engineering discipline~\citep{kaplan2020scaling, hoffmann2022training, bi2024deepseek}. 
Before launching a large-scale training job with a compute budget of $C$ FLOPs, three critical questions need to be answered~\citep{muennighoff2023scaling}: (i) \textbf{(Performance)} What is the best achievable performance? 
% (ii) \textbf{(Allocation)} What are the optimal model and dataset sizes for achieving peak performance? 
(ii) \textbf{(Allocation)} What is the optimal way of allocating the compute budget among axes such as the model size and the dataset size? 
(iii) \textbf{(Hyperparameter)} What are the optimal hyperparameters for the training job?
Scaling laws provide a tool for researchers to answer these questions by conducting inexpensive experiments at smaller scales and extrapolating to larger scales.

In contrast to the success of scaling in (self-)supervised learning, reinforcement learning (RL) has yet to find its own guiding principles for predictive scaling. While scaling has powered landmark achievements from game-playing agents~\citep{silver2017mastering, vinyals2019grandmaster, berner2019dota}, to fine-tuning large language models~\citep{guo2025deepseek, khatri2026art}, the foundational scaling dynamics of RL remain poorly understood. 
Unlike (self-)supervised learning, where the data distribution is stationary, RL agents learn from experiences that continuously change as their own policy evolves. Learning in RL is thus inherently non-stationary. This active coupling between policy improvement and data generation implies that RL may be governed by scaling laws distinct from those of passive, supervised regimes. 
Moreover, unlike the cross entropy loss which has a clear information-theoretical grounding, rewards in RL are often designed in an ad-hoc manner based on heuristics or domain knowledge. Consequently, the common RL performance metric, mean episode return, may not vary smoothly~\citep{hilton2023scaling}. 
Due to these differences, it is unclear if the same conclusion, and predictive power, from the (self-)supervised learning scaling laws apply to RL. 

Existing work on RL scaling laws primarily answer only a subset of the aforementioned three questions while also lacking validation of the predictive performance of the fitted laws~\citep{jones2021scaling, hilton2023scaling, neumann2023scaling}. We argue that the most critical value of scaling laws is in their predictive power and thus validating the predictions is a centerpiece of scaling law studies. 
In this paper, we present an empirical study of scaling laws in RL. We study all three questions with validation of the predictive performance from small-scale experiments to large-scale experiments in each case. 
We conduct experiments in two procedurally generated environments, Starpilot~\citep{cobbe2020leveraging} and Craftax~\citep{pmlr-v235-matthews24a}. We fit scaling laws for the optimal hyperparameters, model and data sizes, and performance as a function of total training FLOPs $C$. We validate the predictive power of the fitted laws by testing their prediction accuracy at one or two orders of magnitude larger training FLOPs. 
To overcome the non-smoothness of the mean episode return, we employ a variant of the Elo score~\citep{elo1978rating} for the single-agent RL paradigm and showed that it extrapolates effectively whereas the mean episode return does not, making it a critical choice for scaling studies.
Our analysis in Craftax reveals that, despite the empirical accuracy of the fitted laws, the compute-optimal frontier does not follow a global log-linear law throughout training. We connect this observation to a phenomenon known as Ray interference~\citep{schaul2019ray}. Based on this observation, we argue against the existence of scaling laws for arbitrary RL environments.

Our main contributions are:
\begin{itemize}
    \item We establish a two-stage empirical framework for studying RL scaling laws. Our experimental results show that optimal model size and key hyperparameters (e.g., batch size and learning rate) follow predictable power-law relationships with respect to training compute.
    \item We address the non-smoothness of mean episode return and show that a single-agent Elo metric scales log-linearly as a function of training compute, enabling reliable performance extrapolation up to two orders of magnitude.
    \item Guided by our scaling laws, we train an actor-critic agent to $1\text{e}{19}$ FLOPs and set a new state-of-the-art performance of $54.0\%$ on the challenging Craftax benchmark.
    \item We identify a local inverse scaling pattern in Craftax and connect it to Ray interference, showing how global trends can mask non-monotonic local dynamics.
\end{itemize}

%% file: related_work.tex
\section{Related Work}
\label{sec:related}

In the past decade, empirical power laws were discovered in multiple deep learning application domains including image classification~\citep{hestness2017deep,rosenfeld2020a}, image generation~\citep{henighan2020scaling}, and language modeling~\citep{kaplan2020scaling, hoffmann2022training}.
~\citet{kaplan2020scaling} observed that the cross entropy loss of an autoregressive language model scales as a power law with the model size, the dataset size, and the amount of compute used for training across several orders of magnitude. Their work provides a method to answer the \textit{performance} question. Based on their fitted scaling laws,~\citet{kaplan2020scaling} recommended scaling model size more aggressively than dataset size as training compute increases.
Following~\citet{kaplan2020scaling}, \citet{hoffmann2022training} conducted a study on the optimal model size for a certain training compute budget, directly addressing the \textit{allocation} question. Contradictory to~\citet{kaplan2020scaling}, their work showed that the model size and dataset size should grow at an equal pace while scaling up training compute. 
In both works, the power laws were fitted to explain the observed data. Their predictions were not validated at larger scales. Moreover, many experimental details such as hyperparameter choices are missing or poorly studied, leaving the \textit{hyperparameter}  question unanswered. To address these limitations, DeepSeek~\citep{bi2024deepseek} studied all three questions of performance, allocation, and hyperparameter and tested the predictive accuracy of the fitted scaling laws at larger scales.
Our work builds upon these works. We employed their general methodology to study all three questions and tested the predictive accuracy of the corresponding scaling laws. 
Meanwhile, our work is different from these works as we focus on RL instead of language modeling.

There are few existing works on RL scaling laws. The most closely related work is by~\citet{hilton2023scaling}, who studied scaling laws of single-agent RL. They identified the difficulty of predicting the mean episode return at larger training compute due to its non-smoothness. To address this challenge, they proposed a new metric called \textit{intrinsic performance} which measures the performance of an RL agent by the minimum amount of training compute needed to achieve parity with it. Although by definition intrinsic performance is not suited for answering the performance question, it served as a tool to answer the allocation question.
~\citet{fu2026compute} studied scaling laws for value-based RL algorithms with a focus on the allocation question and the hyperparameter question. 
~\citet{jones2021scaling} and~\citet{neumann2023scaling} studied scaling laws of the AlphaZero agent~\citep{silver2017mastering} in the context of two-player zero-sum games. Both works found that the Elo~\citep{elo1978rating} of an agent scales predictably with the amount of training compute. \citet{jones2021scaling} further showed that the Elo scaling law itself scales with the size of the game.
Like~\citet{hilton2023scaling}, we focus on the single-agent RL setup. Instead of using intrinsic performance, we employ an Elo variant inspired by~\citet{jones2021scaling, neumann2023scaling} as the performance metric which allows us to answer the performance question. In contrast to prior works, which only studied a subset of the three questions, to the best of our knowledge our work is the first to study all three questions in the context of RL, rendering a more complete picture of RL scaling laws.

Recently, the success of reasoning models~\citep{jaech2024openai, comanici2025gemini, guo2025deepseek} reinvigorated interest in RL scaling laws in the context of large language model (LLM) finetuning. Examples include~\citet{khatri2026art},~\citet{cheng2026isocompute}, and~\citet{tan2026scaling}. These works focus on a specific application domain of RL and concern domain-specific quantities such as the number of parallel rollouts per prompt and the pass rate on a specific dataset. In contrast, our work focuses on the more general sequential-decision making setup with minimum assumptions about the environment and the reward structures.

%% file: method.tex
\section{Method}
\label{sec:method}

Our methodology implements a two-stage framework adapted from the paradigm proposed by~\citet{bi2024deepseek}. First, we characterize hyperparameter scaling dynamics by conducting a systematic grid search across small-to-medium compute budgets to establish empirical power laws that parameterize optimal configurations as functions of training compute. Second, we determine the joint scaling behavior of model size and performance using the isoFLOPs profile approach ~\citep{hoffmann2022training}, dynamically configuring the hyperparameters in this stage using the optimal values prescribed by our first-stage scaling laws. To validate this framework, we first conduct a rigorous scientific evaluation of predictive performance at a $10\times$ compute frontier by independently comparing each scaling law's prediction against the empirical ground truth. We then showcase the practical utility of these scaling laws at a $100\times$ compute scale, mirroring real-world constraints where validating prediction accuracy is computationally prohibitive. Rather than evaluating the laws themselves, this step demonstrates their use in practice: we execute a large-scale training run configured by our hyperparameter and model size prescriptions, comparing its realized performance to the performance scaling law's prediction to verify that the high-budget training job meets expectation.

\subsection{Hyperparameter scaling laws}
We used the A2C algorithm~\citep{pmlr-v48-mniha16} across all experiments in this work. In addition to batch size and learning rate, the entropy regularization strength is also an important hyperparameter in A2C. We did a preliminary grid search over combinations of these three hyperparameters at a low compute regime. The optimal entropy regularization strength is consistent across different compute budgets. Following this observation, we decided to focus on studying the scaling dynamics of batch size and learning rate. 
Specifically, we fixed the rollout length of each trajectory and varied batch size by changing the number of parallel environment instances.

To characterize the scaling dynamics of hyperparameters, we first need to find their optimal values for a set of different training budgets at a small scale. 
For each training FLOPs $C$, we conduct a grid search on batch size and learning rate across a few different model sizes to identify their optimal values. 
Due to randomness in the training process, the empirical performance is a noisy estimate of the true performance. Thus the empirical optimal hyperparameter configuration is often not the true optimum. RL performance evaluation is known to suffer from higher variance than supervised learning~\citep{henderson2018deep}, which makes the identification problem even harder. 
To reduce the variance of the empirical performance measurement, we conduct multiple independent jobs for each configuration with different random seeds. We then aggregate the episode returns across random seeds by their interquartile mean (IQM). 
Moreover, to take into account the randomness of the empirical results, we do bootstrapping by resampling the random seeds for the experiments. For each resampling, we pick the batch size and learning rate that yields the highest IQM of the episode return. We call this configuration a \textit{competitive} configuration. Through bootstrapping, we obtain a set of competitive configurations for each training compute $C$. 
Instead of fitting the empirically optimal configurations, we fit power laws to all competitive configurations in the following forms:
\begin{align}\label{eq:hyper-law}
B^{*}(C) &= a_{B} C^{d_{B}} \\ 
\alpha^{*}(C) &= a_{\alpha} C^{d_{\alpha}}
\end{align}
$B^{*}(C)$ and $\alpha^{*}(C)$ represent the average competitive batch size and learning rate at training FLOPs $C$ respectively. $a_{B}$, $d_{B}$, $a_{\alpha}$, and $d_{\alpha}$ are parameters fitted from the data. 

\citet{bi2024deepseek} employ a method that shares the same spirit as ours to combat the noise in optimal hyperparameter identification. They define ``near-optimal'' hyperparameters as the ones that yield a loss within $0.25\%$ of the empirical minimum. Similar to our method, they then fit the scaling laws to the near-optimal hyperparameters. 
We do not use the same method because unlike cross entropy the episode return need not vary smoothly through training. It is hard to pick a threshold for the ``near-optimal'' definition. Instead, we rely on bootstrapping to address randomness in experiments.

\subsection{Allocation scaling laws}
To understand the optimal allocation of training compute, we first need to calculate the total FLOPs $C$ of a training run. In Transformer-based language models, this is often approximated by $C=6ND$ where $N$ denotes the total number of parameters in the model and $D$ denotes the total number of tokens consumed during training~\citep{kaplan2020scaling,hoffmann2022training}. 
This approximation is based on dense layers. However, we use convolutional networks which do not follow the same calculation. Therefore, similar to the FLOPs-per-token notion used by~\citet{bi2024deepseek}, we use FLOPs-per-frame (FPF), denoted by $M$, as a measurement of the model size. Here a frame refers to the observation of a single step in the environment. FPF is defined as the number of FLOPs spent on each observation during training. The total training compute is thus $C = MD$ where $D$ denotes the number of frames the agent is trained for. Note that for a model-based RL agent we may need to take additional factors such as planning steps into account. We focus on model-free RL agents in this work and leave the study of model-based agents to future work.

With the clear factoring of training FLOPs, we use the isoFLOPs profile approach~\citep{hoffmann2022training} to study the optimal compute allocation between model size and data size. We need to find the optimal allocation at a set of different training FLOPs at small scale. 
For each training FLOPs $C$, we train models of different sizes using hyperparameters predicted by their corresponding scaling laws. To combat variance, we run multiple independent jobs for each model size with different random seeds and aggregate the results by IQM. Then we fit a parabola to approximate the mapping from FPF to performance. We extract the optimum of the parabola to identify the optimal FPF $M^{*}$ for training FLOPs $C$. 
Lastly, we fit a power law in the following form:
\begin{align} \label{eq:fpf-law}
M^{*}(C) = a_{M} C^{d_{M}}
\end{align}
$a_{M}$ and $d_{M}$ are parameters fitted from data. 
We do not need to fit a separate law for the optimal data size $D^{*}$ as it can be derived from $M^{*}$.

\subsection{Performance scaling laws in Elo}
Before predicting the performance as a function of the training compute, we must first determine how performance is measured. 
As observed by~\citet{hilton2023scaling}, the compute-optimal frontier need not vary smoothly when the performance is measured in mean episode return. The fundamental reason is that the magnitude of rewards (and thus returns) does not track learning progress continuously. 
To address this problem, we employ a variant of the Elo score~\citep{elo1978rating} as a performance measurement of a~\textit{single-agent} RL training system.
Elo is typically used to measure the relative playing strength of different players within a league in two-player games. 
In this experiment, the league consists of the final agent checkpoint of each training run. We compute the Elo for each agent in our single-agent environment by creating simulated two-player games. We reserve a fixed set of environment instances for evaluation. Every agent is evaluated on this evaluation set. We then set up a simulated tournament between every pair of agents. For each pair, we treat each evaluation environment as an independent competition where the agent that achieved a higher return wins. If both agents achieve the same score, it is a tie and counts $0.5$ win for each agent. By aggregating over the entire evaluation set, we compute a win rate of one agent against the other. We do this for every pair of agents in the league and obtain a win rate matrix where cell $(i, j)$ denotes the win rate of agent $i$ against agent $j$. By fitting a Bradley-Terry model~\citep{Bradley1952RankAO} to this win rate matrix, we obtain an Elo score for each agent. Note that Elo is a relative metric and thus the solution is not unique. To anchor it, we subtract every agent's Elo by the Elo of the random policy post hoc.

For the performance scaling laws, we use the same data from the allocation experiment. We solve the parabola to find the optimal performance for each training compute $C$ and fitted a scaling law in Elo.
Since the Elo score itself is already in log-scale, instead of a power law, we fit a log-linear law in the form below where $d_{\text{Elo}}$ and $a_{\text{Elo}}$ are parameters fitted from data:
\begin{align} \label{eq:elo-law}
\text{Elo}^{*}(C) = d_{\text{Elo}} \log_{10}{C} + a_{\text{Elo}}
\end{align}

%% file: experiment.tex
\section{Experiment Results}
\label{sec:experiment}

\begin{table}[t]
    \centering
    \begin{tabular}{ccccccccc}
        \hline
        Environment & $a_{B}$ & $d_{B}$ & $a_{\alpha}$ & $d_{\alpha}$ & $a_{M}$ & $d_{M}$ & $a_{\text{Elo}}$ & $d_{\text{Elo}}$ \\
        \hline
        Starpilot & $7.14\text{e}{-4}$ & $0.39$ & $4.08\text{e}{-3}$ & $-0.023$ & $5.3$ & $0.44$ & $-3186$ & $240.97$ \\
        Craftax & $0.28$ & $0.26$ & $1.57\text{e}{-2}$ & $-0.082$ & $0.53$ & $0.43$ & $-1987$ & $223.86$ \\
        \hline
    \end{tabular}
    \caption{Summary of all fitted coefficients and exponents in the experiments presented in Section~\ref{sec:experiment}.}
    \label{tab:coeffs}
\end{table}

In this section we present experiment results in Starpilot~\citep{cobbe2020leveraging} and Craftax~\citep{pmlr-v235-matthews24a}. 
In the following sections we present the fitted scaling laws for hyperparameters, compute-optimal model sizes, and optimal performance in detail.
Table~\ref{tab:coeffs} summarizes the parameters of all fitted laws. 

\subsection{Hyperparameters}
In this experiment, we selected $7$ training budgets between $1\text{e}{14}$ FLOPs and $1\text{e}{17}$ FLOPs. For each budget, we conducted a grid search of $7$ different learning rates and $7$ different batch sizes with $4$ models of different sizes. 
% Full details of the search are presented in the Appendix. 
We used the bootstrapping method described in Section~\ref{sec:method} to identify a set of $1000$ competitive batch sizes and learning rates at each training budget. 
Then we fitted power laws for the learning rate and the batch size using data between $1\text{e}{14}$ FLOPs and $1\text{e}{16}$ FLOPs and evaluated their predictions at $1\text{e}{17}$ FLOPs. 
In Craftax, $1\text{e}{14}$ FLOPs was excluded from the power law fitting because we observed a violation of monotonic model size scaling from $1\text{e}{14}$ to $3\text{e}{14}$, for which we provide a deeper analysis in Section~\ref{sec:inverse}. 

\begin{figure}[t]
	\centering
	\includegraphics[width=0.24\columnwidth]{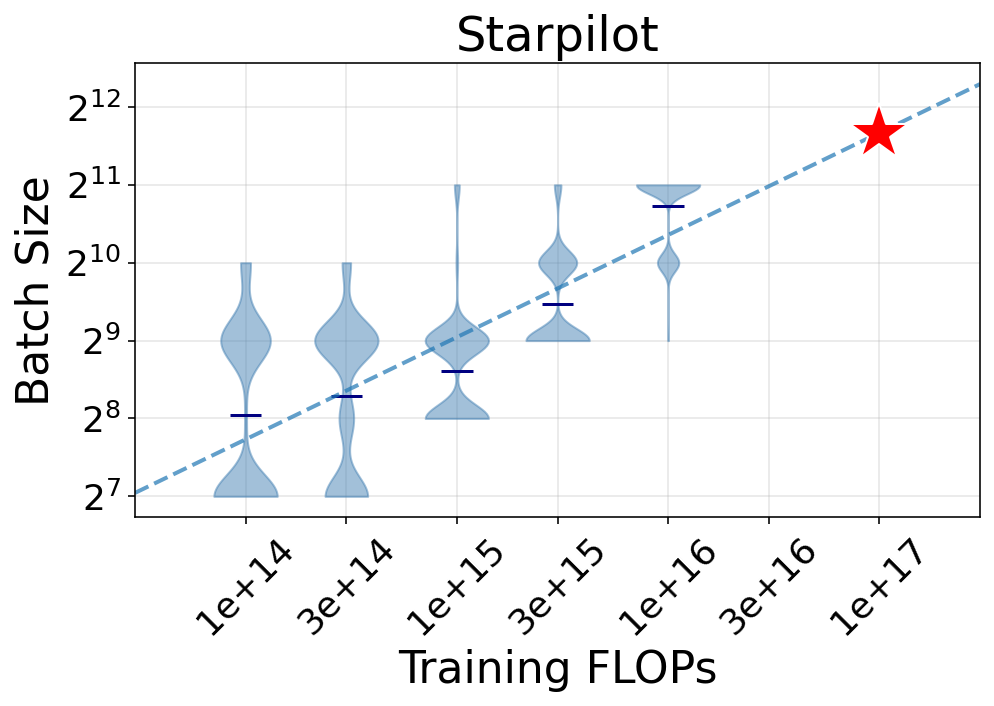}
	\includegraphics[width=0.24\columnwidth]{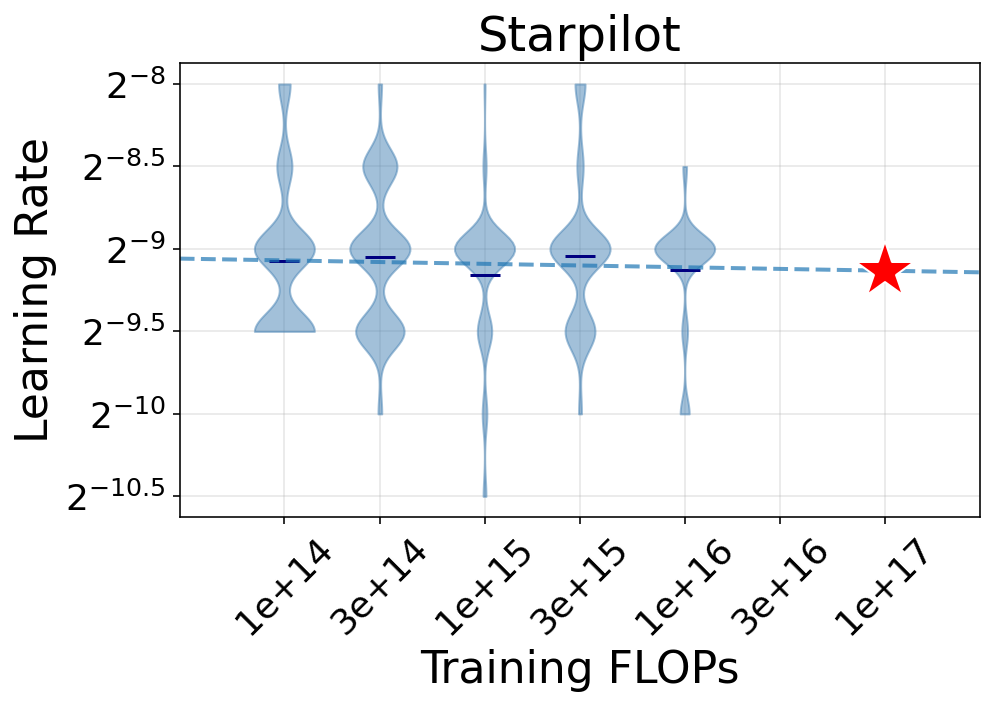}
	\includegraphics[width=0.24\columnwidth]{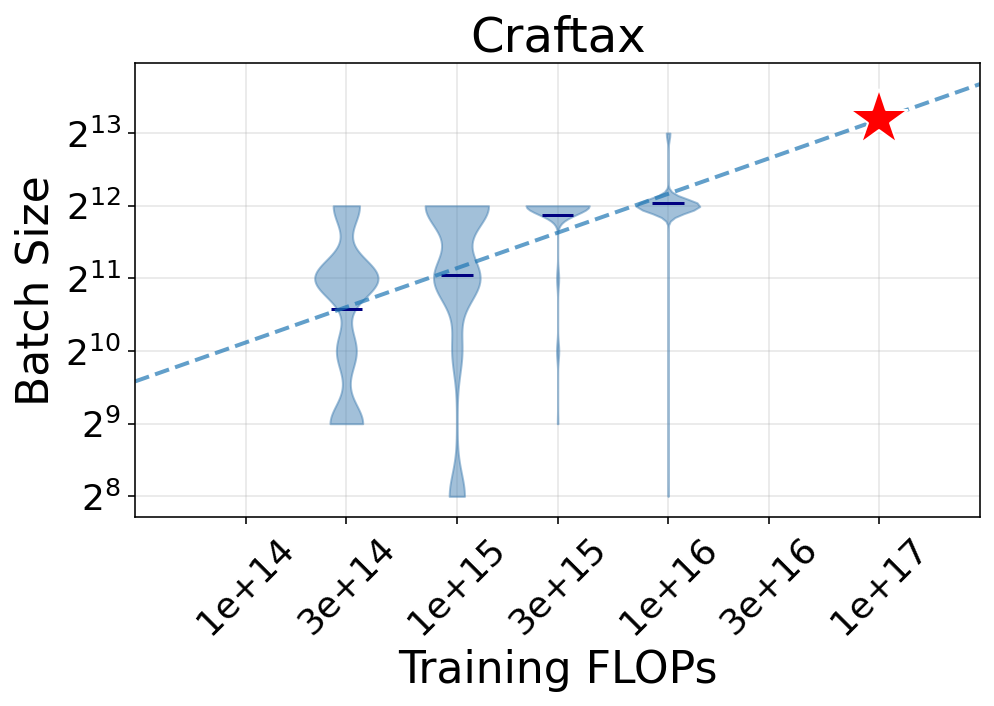}
	\includegraphics[width=0.24\columnwidth]{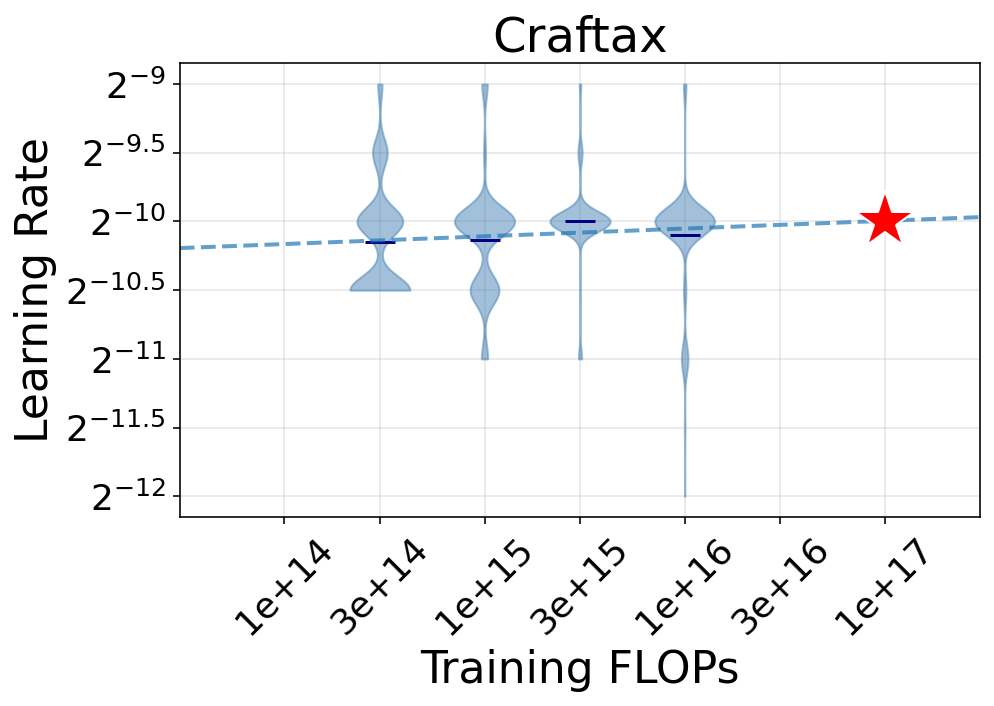}
	\caption{The violin plots show the distributions of the competitive batch sizes and learning rates respectively. The dashed lines visualize the fitted power laws. Red stars mark predicted values.}
    \cutcaptiondown
	\label{fig:hparams_fit}
\end{figure}

\begin{figure}[t]
	\centering
	\includegraphics[width=0.44\columnwidth]{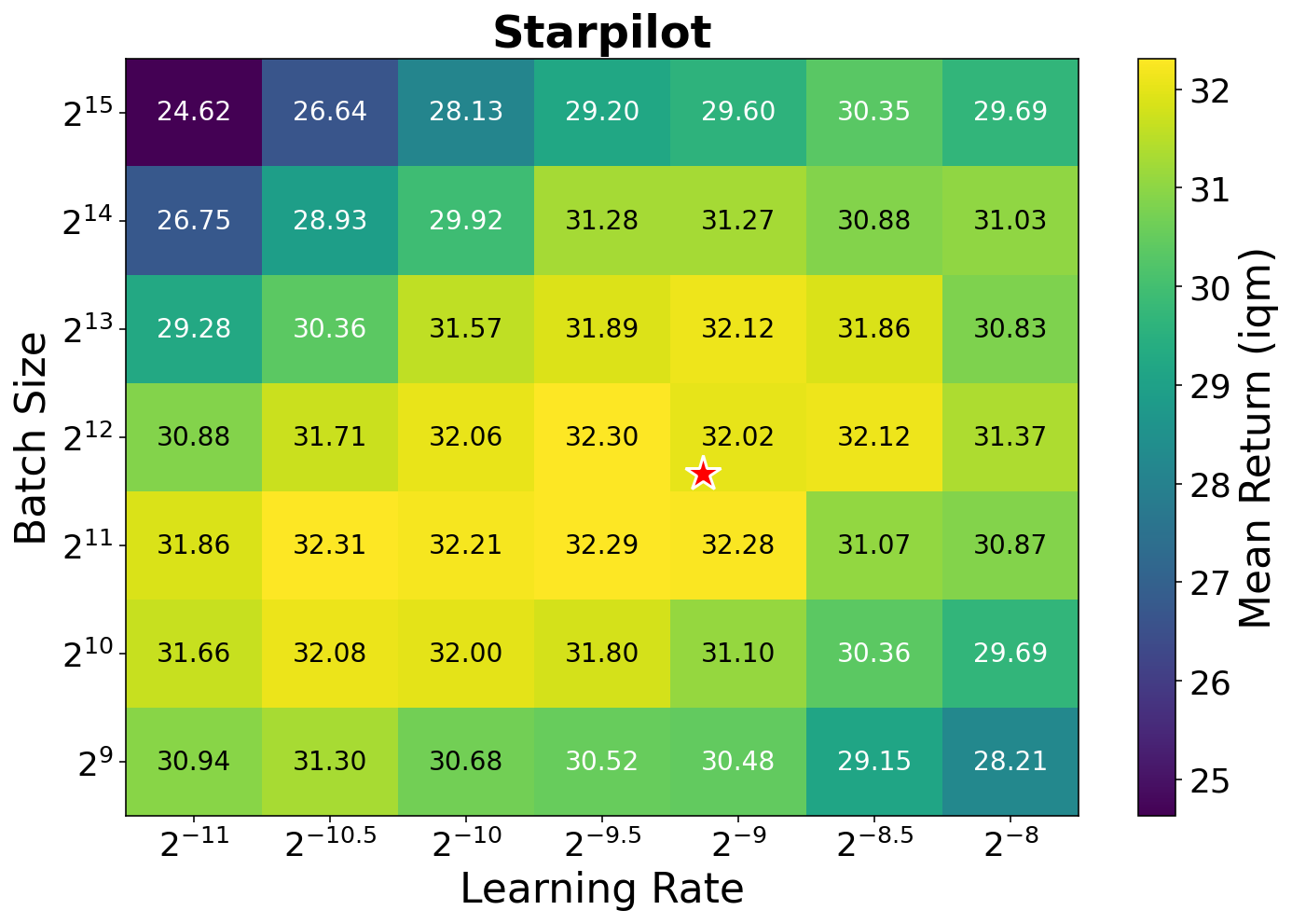}
	\includegraphics[width=0.44\columnwidth]{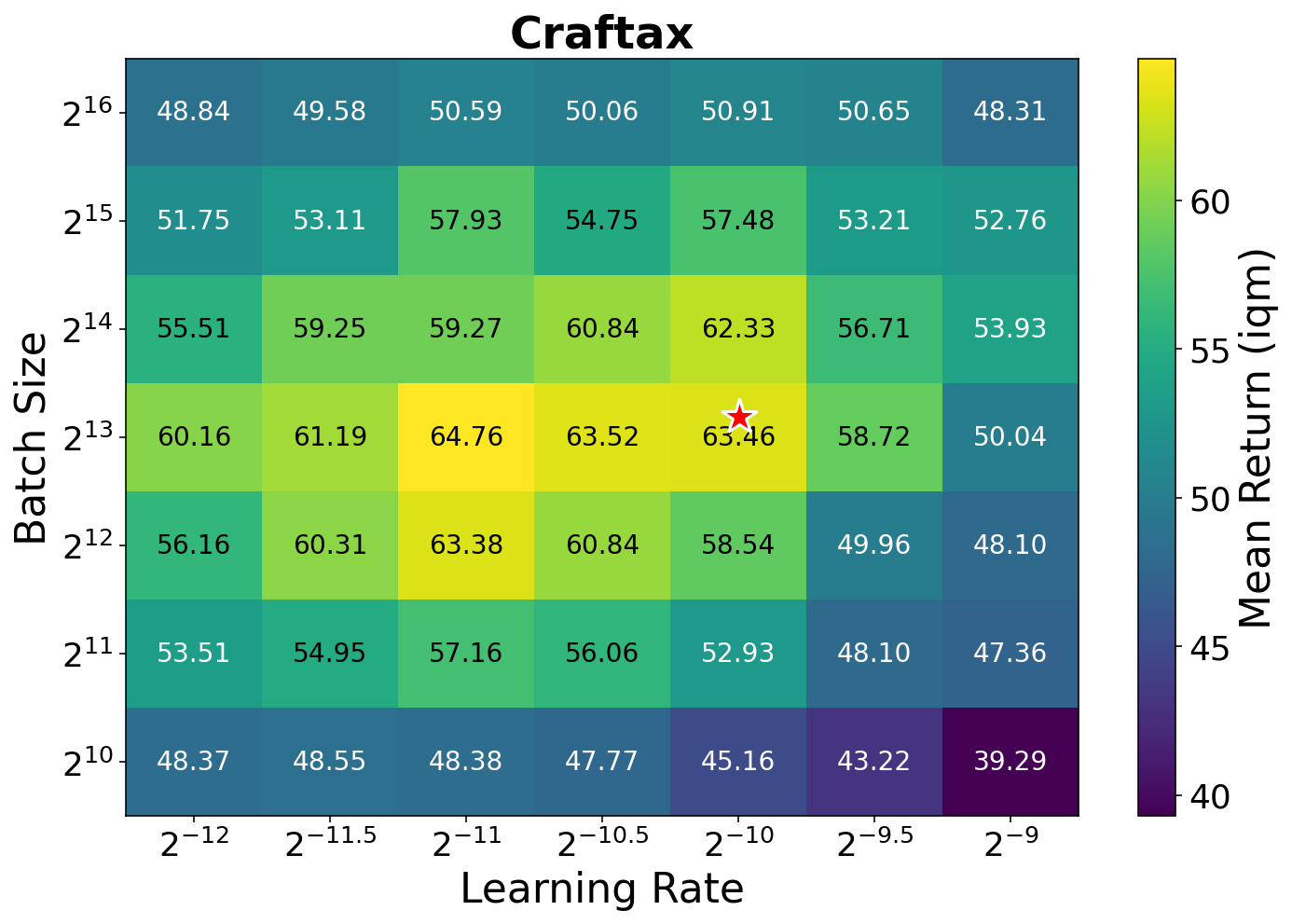}
	\caption{Visualization of performance contour of different hyperparameter configurations at $1\text{e}{17}$ training FLOPs. 
    The number in each cell is the mean episode return on the evaluation set, aggregated by IQM across $5$ seeds. 
    The red stars denote the predicted configurations by the scaling laws.}
	\label{fig:hparams_eval}
    \cutcaptiondown
\end{figure}

Figure~\ref{fig:hparams_fit} visualizes the distributions of the competitive hyperparameters and the fitted power laws. 
In both Starpilot and Craftax, the optimal batch size increases as a function of the training FLOPs. The optimal learning rate roughly remains constant with a small negative exponent in Starpilot and a small positive exponent in Craftax. Note that the learning rate scaling trend contradicts the common wisdom that it should decrease as training compute increases. Nonetheless we will see that empirically its predictions are accurate. 
The heatmaps in Figure~\ref{fig:hparams_eval} visualize where the predicted batch sizes and learning rates are located in the performance contour at $1\text{e}{17}$ training FLOPs. 
In both environments, the predictions are located near the center of the high performance region. 
In Starpilot the optimal configuration yielded a mean episode return of $32.3$ while the predicted configuration yielded $32.02$. In Craftax, the optimal mean episode return is $64.76$ while the predicted configuration yielded $63.46$. 
Overall, the predicted hyperparameters yield performance close to the empirical optimum, validating the predictive performance of the scaling laws.

For the next isoFLOPs profile experiment, we used the scaling laws fitted on all data between $1\text{e}{14}$ FLOPs ($3\text{e}{14}$ for Craftax) and $1\text{e}{17}$ FLOPs. The parameters of the fitted laws are shown in Table~\ref{tab:coeffs}.

\subsection{Allocation} \label{sec:allocation}
We used the isoFLOPs profile approach~\citep{hoffmann2022training} to answer the allocation question. 
We selected $6$ training FLOPs budgets: $1\text{e}{15}$, $3\text{e}{15}$, $1\text{e}{16}$, $3\text{e}{16}$, $1\text{e}{17}$, and $1\text{e}{18}$. 
For each training budget of $C$ FLOPs, we trained multiple different model sizes. We kept the depth of the neural network fixed and varied the width to scale the model size. 
The batch sizes and the learning rates were predicted by the scaling laws fitted in the previous section. 
To obtain robust results, we ran $10$ independent trials with different random seeds for each configuration. 
We kept the final agent of each training job and formed a league of them together with a uniform random policy to compute an Elo score for each agent, using the method described in Section~\ref{sec:method}. For each configuration, we aggregated the Elo scores across all random seeds by IQM. 
We fitted a parabola to the relationship between $\log_{10}{M}$ and the Elo score by weighted least squares where each data point is weighted by the inverse of its variance across random seeds. Then we extracted its optima to find the corresponding optimal FPF and optimal performance.

\begin{figure}[t]
	\centering
	\includegraphics[width=0.45\columnwidth]{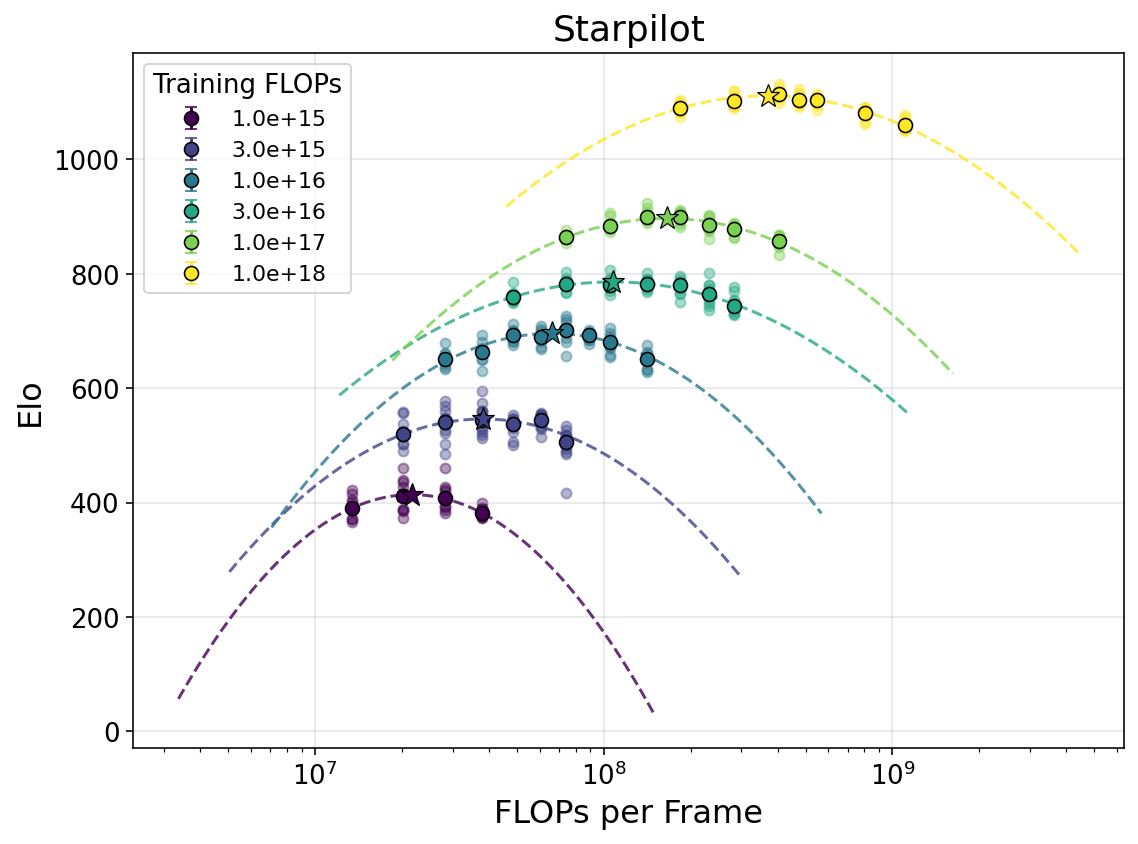}
	\includegraphics[width=0.45\columnwidth]{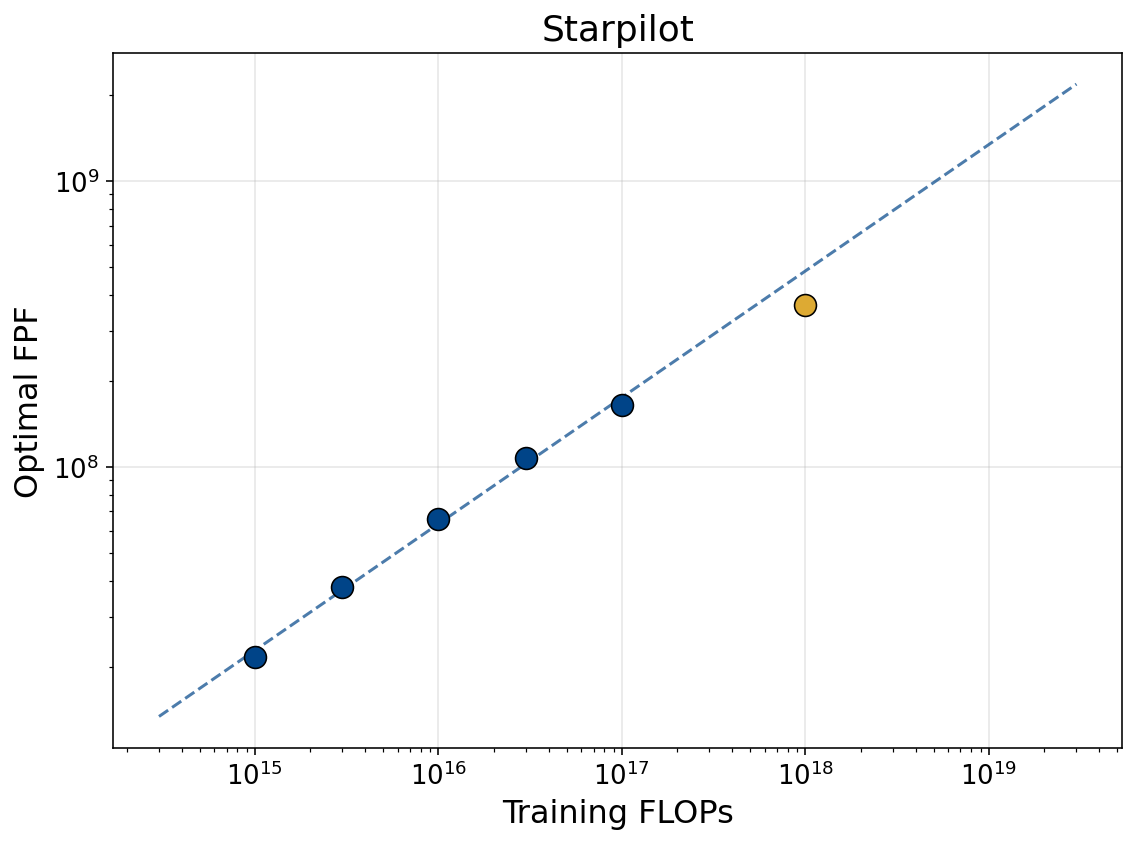}
	\includegraphics[width=0.45\columnwidth]{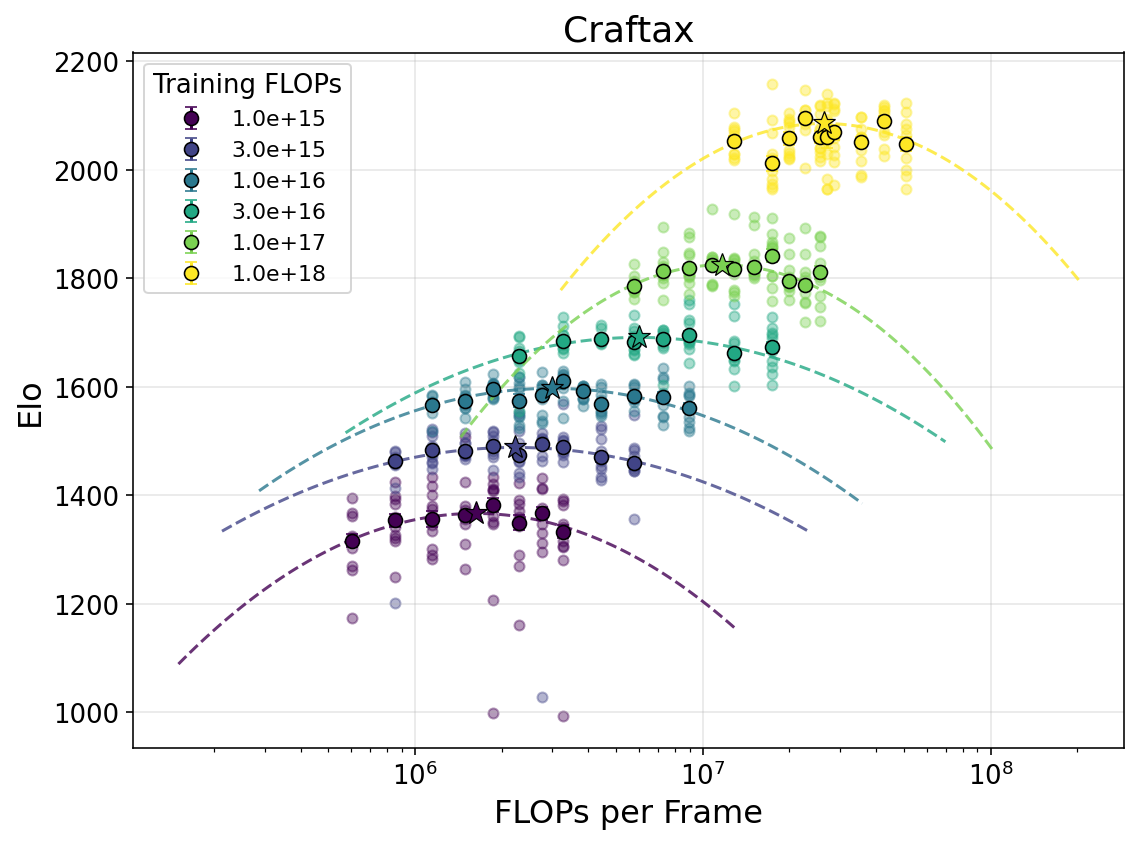}
	\includegraphics[width=0.45\columnwidth]{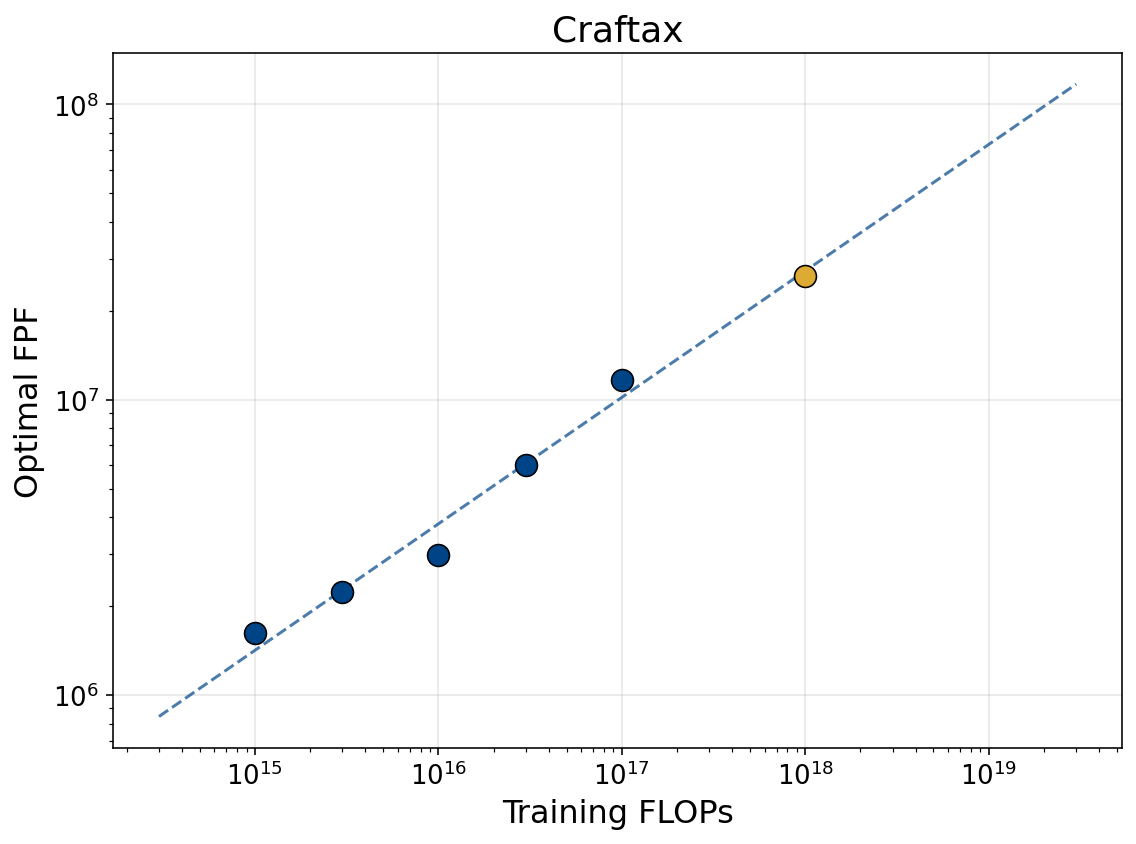}
	\caption{Visualization of data from the isoFLOPs profile experiment in Section~\ref{sec:allocation}. The left column visualizes the final performance of each training job and the fitted parabolas. The x-axis denotes the model size measured in FPF. The y-axis denotes the performance measured in Elo. Each parabola corresponds to one training FLOPs. The right column visualizes the derived optimal FPF and the fitted power law. The x-axis is the training FLOPs and the y-axis is the optimal FPF for each training budget. The blue dots are used for fitting the power laws and the yellow dots are used for evaluation.}
	\label{fig:isoflops}
\end{figure}

The left column in Figure~\ref{fig:isoflops} shows the data points and the fitted parabolas and their optima. 
We fitted a power law in the form of Eq.~\ref{eq:fpf-law} using data between $1\text{e}{15}$ and $1\text{e}{17}$ FLOPs and evaluated its prediction at $1\text{e}{18}$ FLOPs. The right column in Figure~\ref{fig:isoflops} visualizes the fitted laws. The coefficients and exponents of the fitted laws can be found in Table~\ref{tab:coeffs}. The predicted and the empirically optimal FPFs are summarized in Table~\ref{tab:isoflops}.
In Starpilot, the power law fits the data well. Although the prediction overestimates the optimal FPF by $31\%$, the performance of the predicted model size is close to optimal (see the yellow circles and stars at $1\text{e}{18}$ FLOPs in Figure~\ref{fig:performance}). 
In Craftax, the fit is less clean. But nevertheless, the FPF prediction is extremely accurate with a relative error of only $3\%$.
Overall the fitted scaling laws demonstrate strong prediction accuracy in both environments, rendering them reliable for informing the model size decisions for large scale training runs.

\begin{table}[t]
    \centering
    \begin{tabular}{ccccccc}
        \hline
        Environment & \makecell{predicted\\FPF} & \makecell{optimal\\FPF} & \makecell{relative\\error} & \makecell{Elo of\\predicted FPF} & \makecell{optimal Elo} & \makecell{win rate\\error} \\
        \hline
        Starpilot & $4.85\text{e}8$ & $3.69\text{e}8$ & $+31\%$ & $1102.6$ & $1110.7$ & $-1.17\%$ \\
        Craftax & $2.73\text{e}7$ & $2.65\text{e}7$ & $+3\%$ & $2062.8$ & $2087.5$ & $-3.55\%$ \\
        \hline
    \end{tabular}
    \caption{Data for evaluating the prediction accuracy of the optimal model size. We compare the predicted FPF and its realized Elo to the empirical optimum.}
    \label{tab:isoflops}
\end{table}

\subsection{Performance}
\begin{figure}[t]
	\centering
	\includegraphics[width=0.45\columnwidth]{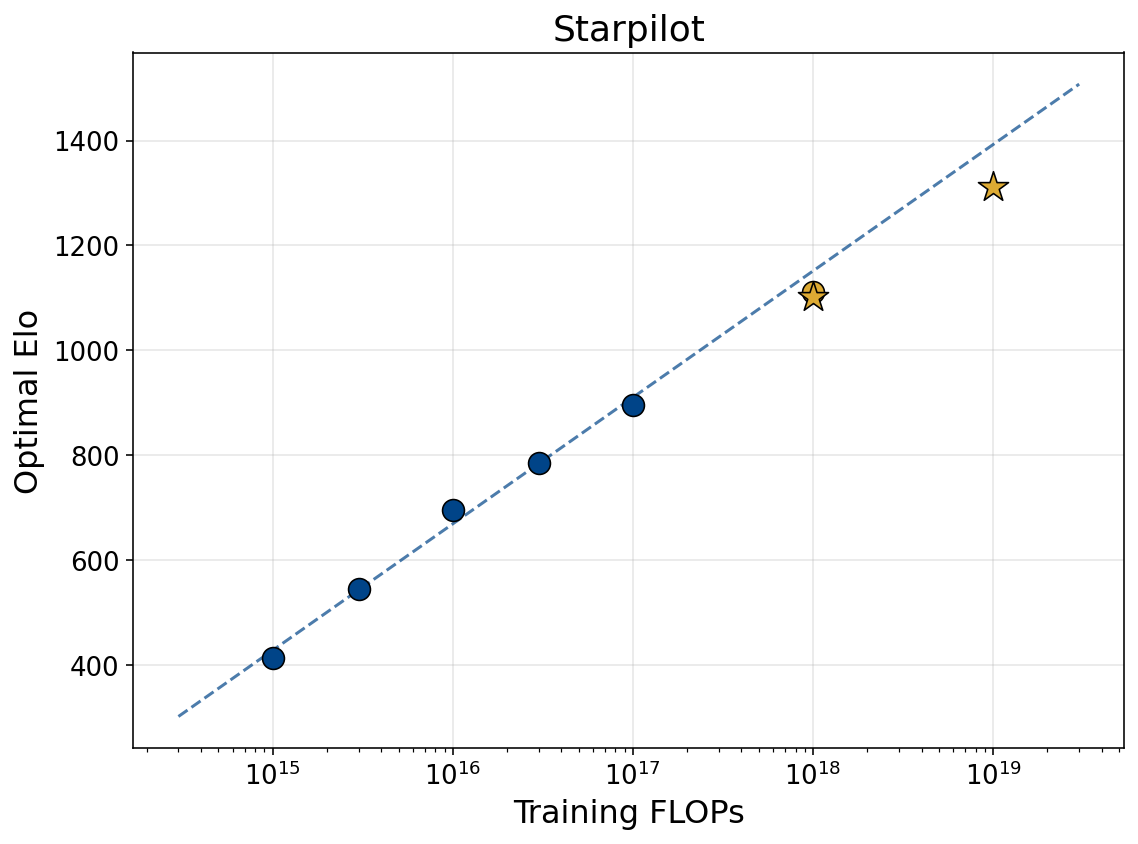}
	\includegraphics[width=0.45\columnwidth]{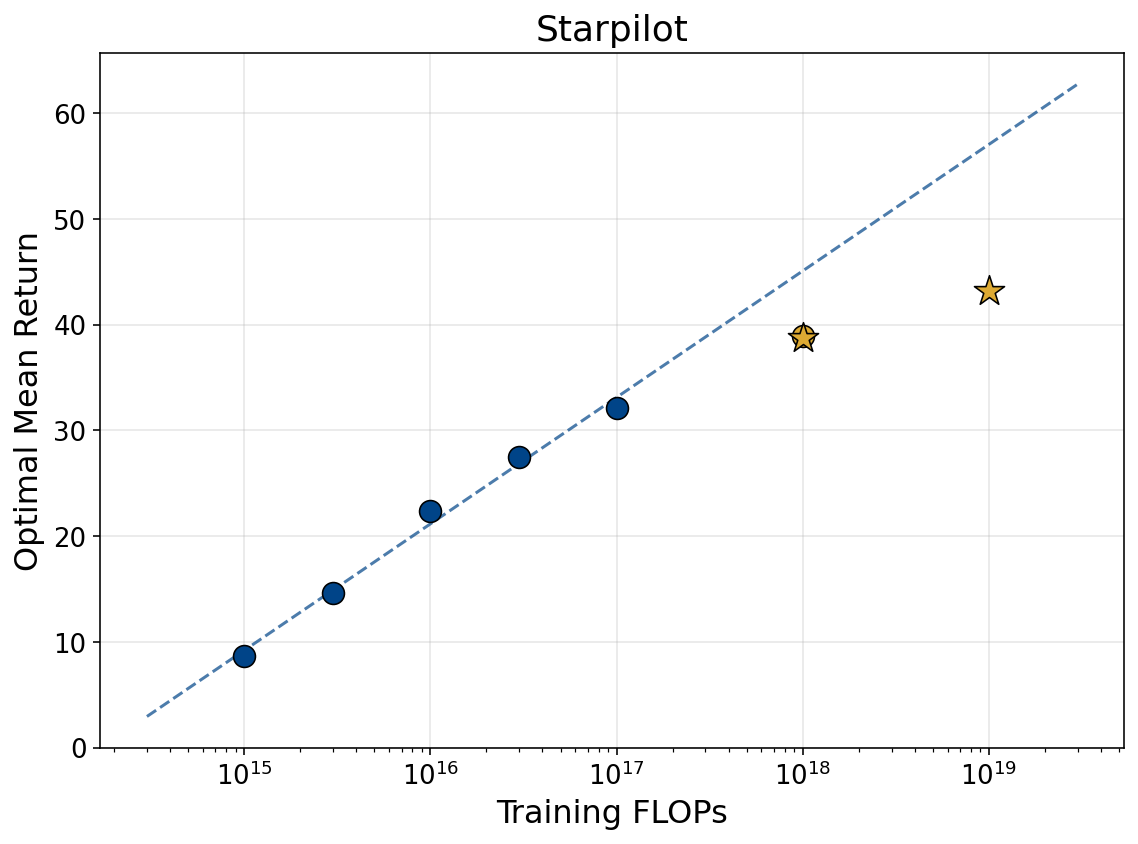}
	\includegraphics[width=0.45\columnwidth]{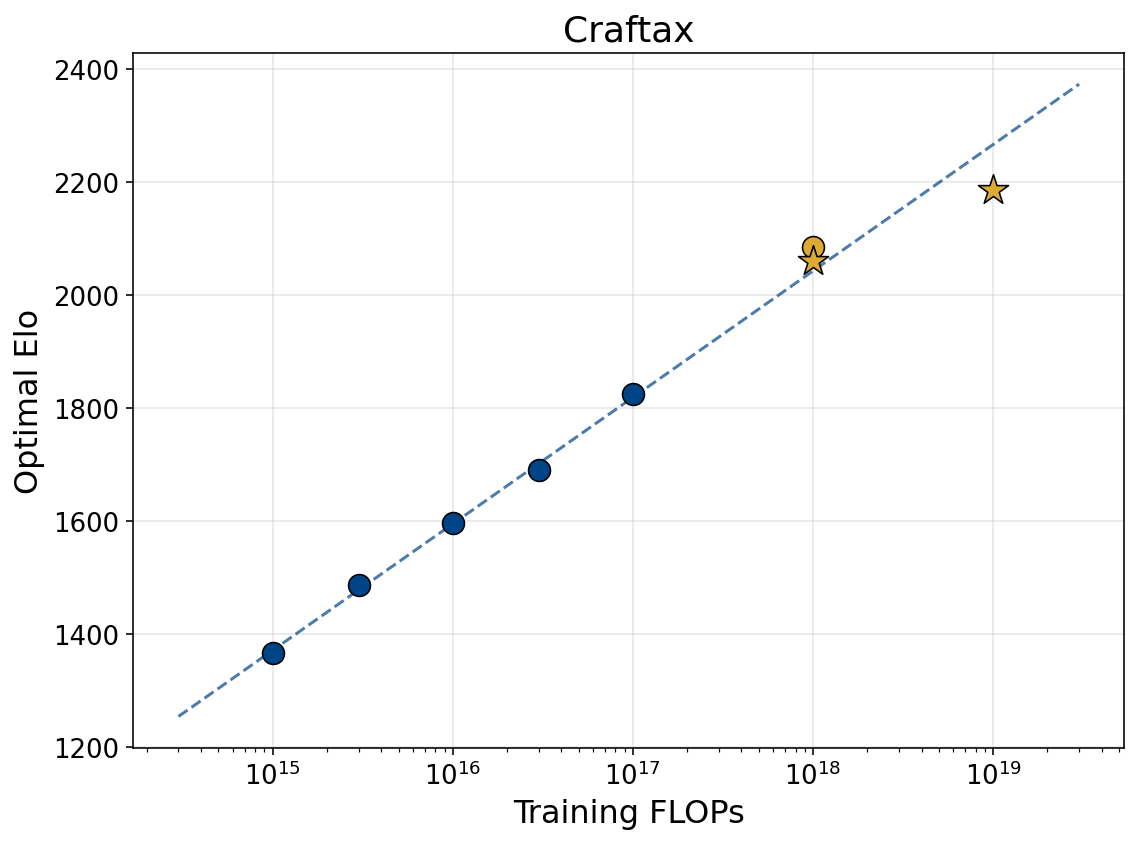}
	\includegraphics[width=0.45\columnwidth]{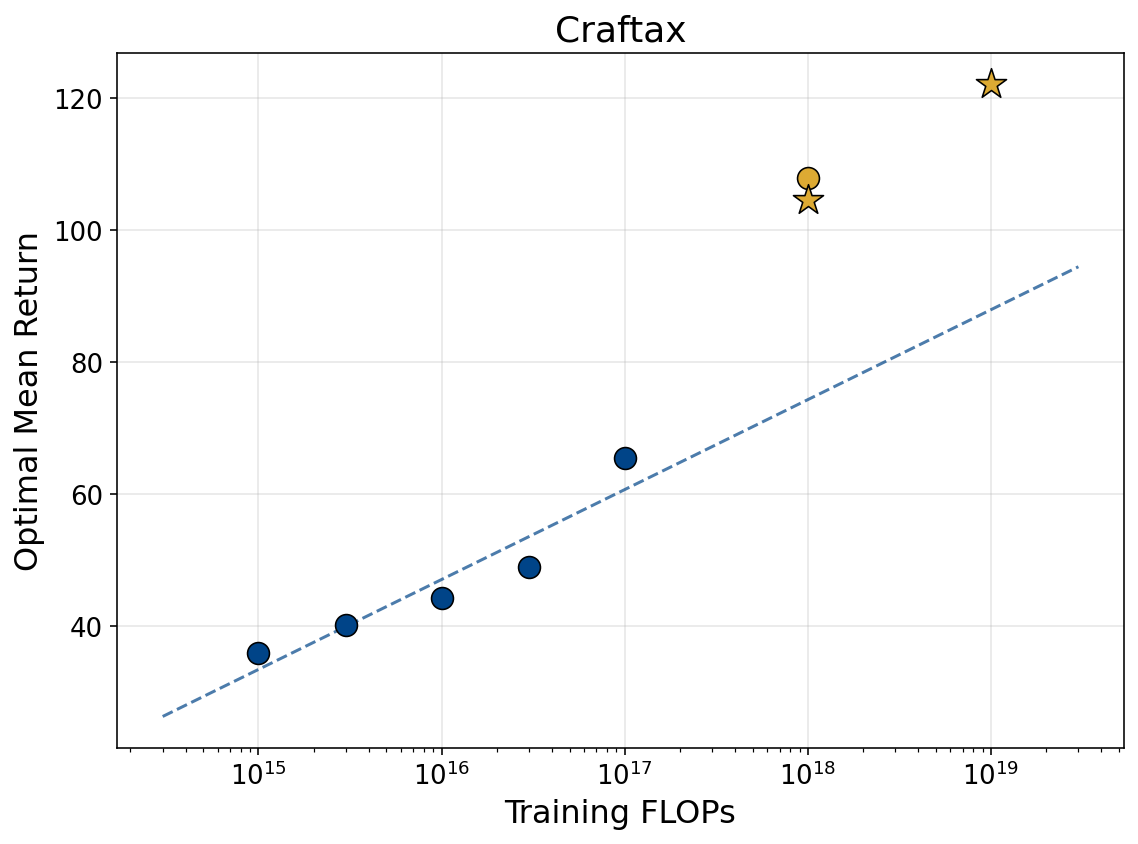}
	\caption{Comparison of different performance metrics. The left column is Elo and the right column is mean episode return. Dots denote empirical optimum. Stars denote the performance of the configuration predicted by the scaling laws. Blue dots are used for fitting the scaling laws. Yellow dots and stars are used for evaluation.}
	\label{fig:performance}
\end{figure}

We used the same data from the previous section to fit a law between $C$ and the Elo score in the form of Eq.~\ref{eq:elo-law}. The parameters of the fitted laws are in Table~\ref{tab:coeffs}. 
The left column in Figure~\ref{fig:performance} visualizes the fitted laws. In both Starpilot and Craftax, the laws fit the data in $[1\text{e}{15}, 1\text{e}{17}]$ well and predict the Elo at $1\text{e}{18}$ accurately, resulting in a $+5.84\%$ and $-6.37\%$ error in win rates respectively.

We also compare Elo to mean episode return as the performance metric. We fitted similar log-linear laws for the mean episode return, which are shown in the right column of Figure~\ref{fig:performance}. In Starpilot, the mean episode return law fits the data well but extrapolates poorly in comparison to the Elo law. In Craftax, the mean episode return law does not even fit the data well and the prediction error at $1\text{e}{18}$ FLOPs is huge. In both environments, Elo extrapolates significantly better than mean episode return, proving itself as the better performance metric for fitting predictive scaling laws.

In each environment, we also conducted a hero run which trained a model with the size predicted by the FPF scaling law for $1\text{e}{19}$ FLOPs. Each run consists of $10$ independent trials with different random seeds. 
The performance results are summarized in Table~\ref{tab:elo} and Figure~\ref{fig:performance}. 
We compare the predicted Elo to the Elo actually achieved. 
% The Elo laws accurately predict the Elo of larger-scale training jobs for two orders of magnitude more FLOPs. 
The deviations are $+80.2$ and $+80.9$ for Starpilot and Craftax respectively, which translate to $+11.34\%$ and $+11.44\%$ deviation in the win rate. 
We want to highlight the result in Craftax. Through a compute-optimal allocation of training resources, we achieve a new state-of-the-art performance of $54.0\%$, more than doubling the previous best published baseline of $20.6\%$ ~\citep{hamon:hal-04659863v1}. 
Note that our hero run consumed significantly more data than the baseline ($135$ billion frames \textit{vs} $4$ billion frames). 
This is intentional as the focus of this work is on compute optimality rather than data efficiency.

\begin{table}[t]
    \centering
    \begin{tabular}{ccccccc}
        \hline
        Environment & \makecell{predicted\\@$1\text{e}{18}$} & \makecell{optimal\\@$1\text{e}{18}$} & \makecell{win rate error\\@$1\text{e}{18}$} & \makecell{predicted\\@$1\text{e}{19}$} & \makecell{actual\\@$1\text{e}{19}$} & \makecell{win rate error\\@$1\text{e}{19}$} \\
        \hline
        Starpilot & $1151.5$ & $1110.7$ & $+5.84\%$ & $1392.5$ & $1312.3$ & $+11.34\%$ \\
        Craftax & $2043.0$ & $2087.5$ & $-6.37\%$ & $2266.8$ & $2185.9$ & $+11.44\%$ \\
        \hline
    \end{tabular}
    \caption{Data for evaluating the accuracy of the Elo predictions. 
    At $1\text{e}{18}$ FLOPs, we compare the predicted Elo to the empirically optimal Elo. 
    At $1\text{e}{19}$ FLOPs, we compare the predicted Elo to the Elo realized by the predicted optimal configuration.
    }
    \label{tab:elo}
\end{table}

\subsection{Inverse scaling analysis} \label{sec:inverse}
\begin{figure}[t]
	\centering
	\includegraphics[width=0.45\columnwidth]{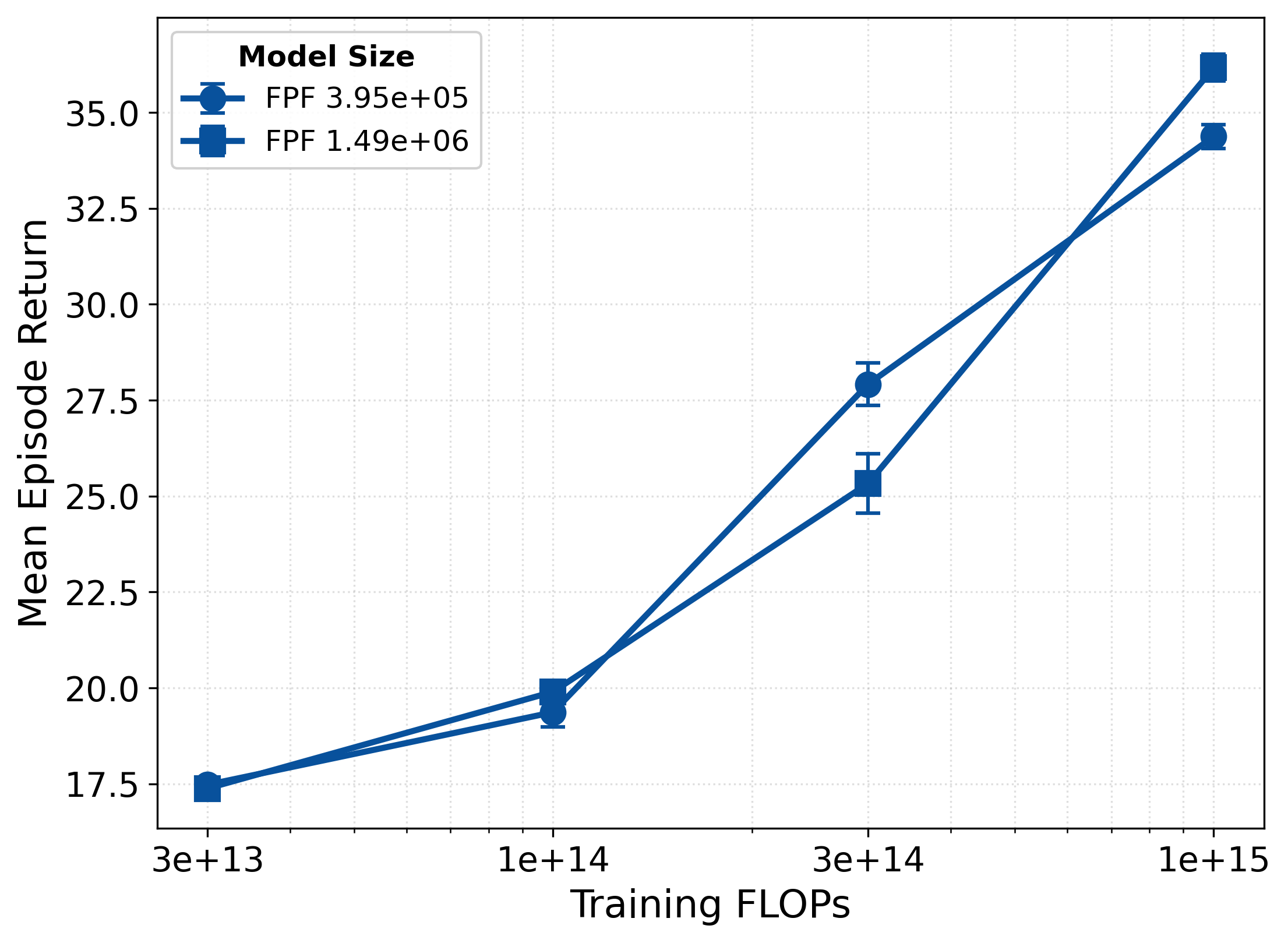}
	\includegraphics[width=0.45\columnwidth]{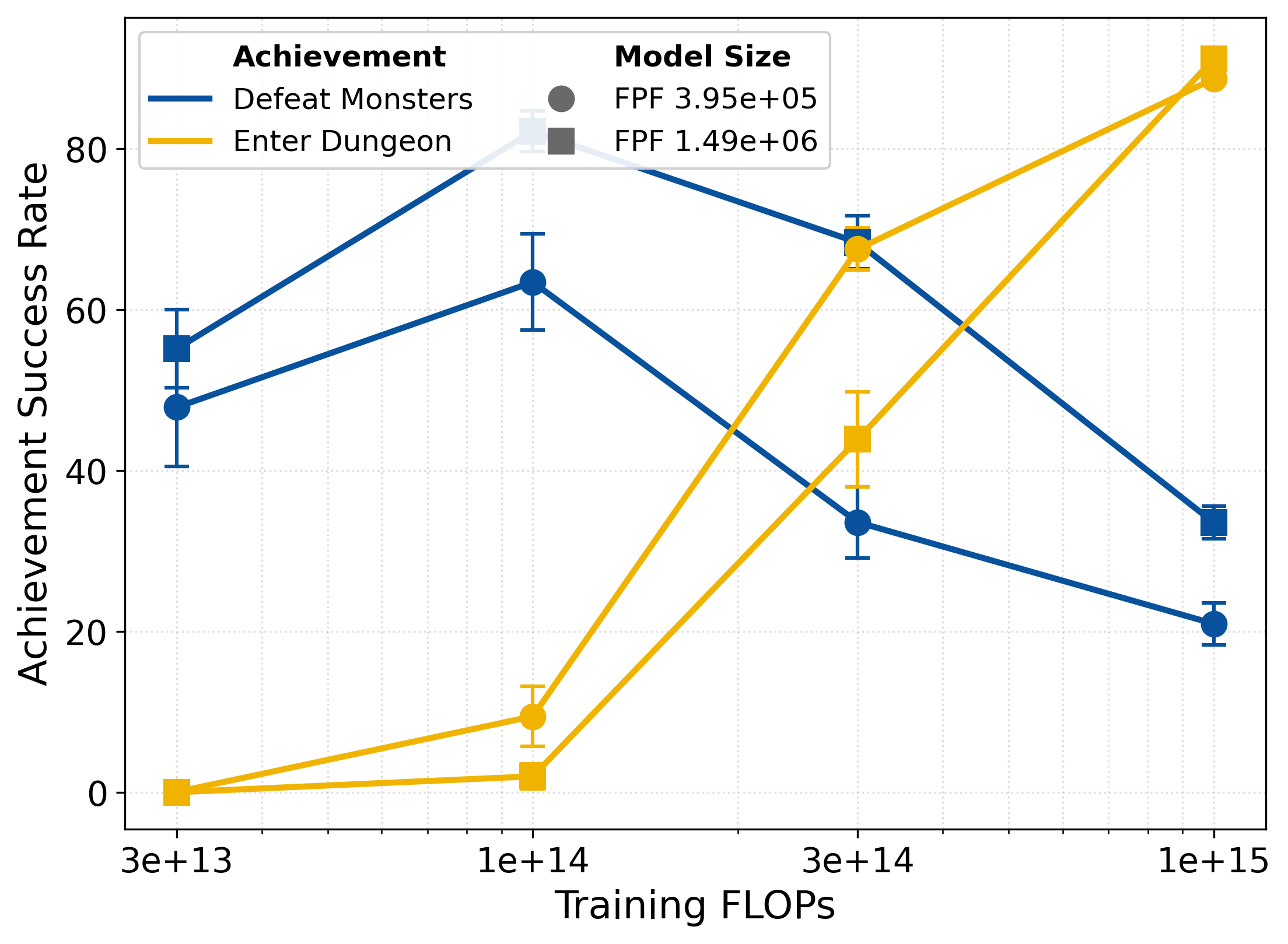}
	\caption{Visualization of the inverse scaling pattern in Craftax. The left plot shows the best performance of two different model sizes across different training FLOPs. 
    The right plot shows the success rate of the enemy defeat achievements and the entering dungeon achievement. 
    Error bars show the standard error across $5$ random seeds.
    }
	\label{fig:craftax_crossover}
    \cutcaptiondown
\end{figure}

In Craftax, we observed a double-crossover between $3\text{e}{13}$ and $1\text{e}{15}$ FLOPs, demonstrating a local inverse scaling pattern where a smaller model outperforms a larger model as the training budget increases. The left plot in Figure~\ref{fig:craftax_crossover} shows this phenomenon. The bigger model outperforms the smaller model at $1\text{e}{14}$ FLOPs but underperforms at $3\text{e}{14}$ FLOPs. Eventually at $1\text{e}{15}$ the bigger model performs better again. This observation contradicts the common belief that the optimal model size grows monotonically as we scale up training compute. 

We conjecture that this phenomenon stems from a specific design of the game dynamics and the pathological learning dynamics it incurs. 
In Craftax, the agent must defeat at least $8$ enemies on each level to unlock the door to the next level. This rule applies to every level except the first, on which the door to the second level (the ``dungeon'') is always open from the beginning of the episode, rendering battling optional. 
However, defeating enemies is relatively easier than entering the dungeon because the door is always located at the center of a ``mountain'', surrounded by stones. Consequently, the agent learns to defeat enemies before switching to entering the dungeon. During the transition, it stops defeating enemies because it is unnecessary though rewarding. In other words, as the experience accumulates, the agent learns to master the more rewarding skill of entering the dungeon while forgetting the less rewarding skill of battling. These learning dynamics differ drastically from the supervised learning setting where the model needs to fit a growing set of data, which naturally demands more capacity. In this case, however, the agent can reallocate the model capacity from battling to path finding without growing the model size. The smaller model then has an advantage over the larger model because it sees more data under the same compute budget. 

Our conjecture is supported by the achievements the agent unlocked during training, shown in the right plot of Figure~\ref{fig:craftax_crossover}. Up to $1\text{e}{14}$ FLOPs, the agent has little success in entering the dungeon regardless of its model size. Progress mainly comes from better battling skills. However, at $3\text{e}{14}$ FLOPs, the agent learns to enter the dungeon while avoiding battles at the same time. Instead of growing its skill set, which demands more model capacity, the agent switches from one skill to another, giving the smaller model advantages. 

This pathology of one reward source suppressing the learning signal from another reward source during the early phase of training resembles Ray interference~\citep{schaul2019ray}. It is a product of the interaction between the environment dynamics and the learning dynamics. Therefore, a different learning algorithm than actor-critic may demonstrate a different scaling behavior. We leave the comparison of the scaling properties of different algorithms to future work.

%% file: conclusion.tex
\section{Conclusion}
\label{sec:conclusion}

In this work, we presented a systematic empirical study of scaling laws in RL, investigating hyperparameter configuration, compute allocation, and performance prediction across the Starpilot and Craftax environments. 
We established that optimal hyperparameters follow power-law relationships with respect to total training compute. 
Furthermore, we also characterized the optimal allocation between model size and data size as functions of training FLOPs. 
To address the non-smoothness of mean episode return, we employed a single-agent variant of the Elo score. We demonstrated that the Elo score scales predictably, enabling robust performance extrapolation up to two orders of magnitude in training compute. Guided by these scaling laws, we scaled up a simple actor-critic algorithm to $1\text{e}{19}$ training FLOPs and achieved a new state-of-the-art performance of $54.0\%$ in the challenging Craftax benchmark. However, our analysis also uncovered a local inverse scaling pathology in Craftax, which we connected to a pathology known as Ray interference, illustrating that global scaling trends can obscure non-monotonic local dynamics.

Despite these contributions, several limitations remain to be addressed in future work. 
First, our empirical study was restricted to a single model-free actor-critic algorithm (A2C); further research is required to verify if our findings generalize to other RL algorithms such as value-based or model-based algorithms. 
Second, while the Elo score offers predictable mathematical scaling, it is fundamentally a relative metric within a closed league of evaluated agents, lacking direct semantic grounding to the underlying task. Consequently, in domains where rewards carry critical semantic meanings, Elo-based scaling laws may be less effective at providing informative or absolute performance predictions.
Finally, our discovery of Ray interference in Craftax indicates that RL scaling is highly sensitive to the specific alignment of environmental dynamics and reward structures, suggesting that a single, universal scaling law may not apply across all arbitrary RL environments. Designing algorithms resilient to Ray interference remains a critical avenue for future research.

%% file: appendix.tex
\section{Training Algorithm Details}
In both Starpilot and Craftax, we train agents using the A2C~\citep{pmlr-v48-mniha16} algorithm. 

\subsection{Neural Network Architectures.}
In Starpilot experiments, we use the IMPALA~\citep{espeholt2018impala} residual convolutional neural network architecture~\citep{he2016deep} with a shared torso between the policy and the value function. Layer normalization~\citep{ba2016layer} is applied between every layer and its following activation function. We vary the model size by the number of convolutional channels of the first layer. The width of subsequent layers scale proportionally.

In Craftax we use the symbolic observations. We apply a $1\times1$ convolutional layer with $E$ channels to the map grid. We then flatten the grid and apply a linear layer with dimension $H=16E$ to compress the map to a compact vector representation. The inventory is encoded to a vector of the same dimensionality $H$ by a linear layer. We fuse the map and the inventory by adding their vectors together. Note that we apply layer normalization to their vectors independently \textit{before} fusion. The fused vector is fed into an LSTM~\citep{hochreiter1997long} cell with $H$ hidden units to handle partial observability. Lastly, we apply linear layers to the output of the LSTM cell to map the state representation to the policy logits and the value function. We used GELU~\citep{Hendrycks2016GaussianEL} activation. Similarly, we apply layer normalization before every activation layer. We control the model size by varying the number of channels in the $1\times1$ convolutional layer $E$.

\subsection{Hyperparameters}
In Starpilot experiments, we use a discount factor of $0.999$.
We set the rollout length to a constant $16$.
The weight for the value function loss is $0.5$ and the entropy regularization strength is $0.01$.
We use the Adam~\citep{kingma2015adam} optimizer with $\beta_{1}=0.9$, $\beta_{2}=0.999$, and $\epsilon=1\text{e}{-8}$. The learning rate is linearly decayed to $0$ through training. We clip the gradient by global norm to $0.5$.

In Craftax, we use a discount factor of $0.999$.
We set the rollout length to a constant $16$. We break each batch of rollouts into $4$ minibatches and thus apply $4$ gradient updates per batch. We do minibatching to improve the overall throughput of the training system because we can employ more parallel environments. The batch size in Section~\ref{sec:experiment} is the number of observations in each minibatch. We apply VTrace~\citep{espeholt2018impala} to handle off-policyness caused by minibatch.
The weight for the value function loss is $0.5$ and the entropy regularization strength is $0.001$.
We use the Adam~\citep{kingma2015adam} optimizer with $\beta_{1}=0.9$, $\beta_{2}=0.99$, and $\epsilon=1\text{e}{-8}$. The learning rate is linearly decayed to $0$ through training. We clip the gradient by global norm to $0.5$.

In both environments, we reserved $1024$ instances for evaluation.

\section{Experiment Configurations}
Table~\ref{tab:starpilot_hparam_sweep} and Table~\ref{tab:craftax_hparam_sweep} summarize the configuration of the grid search in the hyperparameter scaling law experiments for Starpilot and Craftax respectively. 
Table~\ref{tab:isoflops_procgen} and Table~\ref{tab:isoflops_craftax} summarize the configuration for the isoFLOPs profile experiments in Starpilot and Craftax respectively. Note that in the isoFLOPs profile experiments we rounded the batch size to the closest power of $2$ and scaled the paired learning rate by the square-root rule~\citep{malladi2022sdes}.

\begin{table}[ht]
\centering
\small
\begin{tabular}{llll}
\toprule
FLOPs & Model width & Batch size & Learning rate \\
\midrule
$1\text{e}{14}$ & $\{1, 2, 3, 4\}$ & $\{2^{6}, 2^{7}, 2^{8}, 2^{9}, 2^{10}, 2^{11}, 2^{12}\}$ & $\{2^{-11}, 2^{-10.5}, 2^{-10}, 2^{-9.5}, 2^{-9}, 2^{-8.5}, 2^{-8}\}$ \\
$3\text{e}{14}$ & $\{2, 3, 4, 6\}$ & $\{2^{6}, 2^{7}, 2^{8}, 2^{9}, 2^{10}, 2^{11}, 2^{12}\}$ & $\{2^{-11}, 2^{-10.5}, 2^{-10}, 2^{-9.5}, 2^{-9}, 2^{-8.5}, 2^{-8}\}$ \\
$1\text{e}{15}$ & $\{3, 4, 6, 8\}$ & $\{2^{7}, 2^{8}, 2^{9}, 2^{10}, 2^{11}, 2^{12}, 2^{13}\}$ & $\{2^{-11}, 2^{-10.5}, 2^{-10}, 2^{-9.5}, 2^{-9}, 2^{-8.5}, 2^{-8}\}$ \\
$3\text{e}{15}$ & $\{4, 6, 8, 12\}$ & $\{2^{7}, 2^{8}, 2^{9}, 2^{10}, 2^{11}, 2^{12}, 2^{13}\}$ & $\{2^{-11}, 2^{-10.5}, 2^{-10}, 2^{-9.5}, 2^{-9}, 2^{-8.5}, 2^{-8}\}$ \\
$1\text{e}{16}$ & $\{6, 8, 12, 16\}$ & $\{2^{7}, 2^{8}, 2^{9}, 2^{10}, 2^{11}, 2^{12}, 2^{13}\}$ & $\{2^{-11}, 2^{-10.5}, 2^{-10}, 2^{-9.5}, 2^{-9}, 2^{-8.5}, 2^{-8}\}$ \\
$3\text{e}{16}$ & $\{8, 12, 16, 24\}$ & $\{2^{8}, 2^{9}, 2^{10}, 2^{11}, 2^{12}, 2^{13}, 2^{14}\}$ & $\{2^{-11}, 2^{-10.5}, 2^{-10}, 2^{-9.5}, 2^{-9}, 2^{-8.5}, 2^{-8}\}$ \\
$1\text{e}{17}$ & $\{8, 12, 16, 24\}$ & $\{2^{9}, 2^{10}, 2^{11}, 2^{12}, 2^{13}, 2^{14}, 2^{15}\}$ & $\{2^{-11}, 2^{-10.5}, 2^{-10}, 2^{-9.5}, 2^{-9}, 2^{-8.5}, 2^{-8}\}$ \\
\bottomrule
\end{tabular}
\caption{Grid search configurations for fitting hyperparameter scaling laws in Starpilot.}
\label{tab:starpilot_hparam_sweep}
\end{table}

\begin{table}[ht]
\centering
\small
\begin{tabular}{llll}
\toprule
FLOPs & Model width & Batch size & Learning rate \\
\midrule
$3\text{e}{14}$ & $\{3, 4, 6, 8\}$ & $\{2^{7}, 2^{8}, 2^{9}, 2^{10}, 2^{11}, 2^{12}, 2^{13}\}$ & $\{2^{-12}, 2^{-11.5}, 2^{-11}, 2^{-10.5}, 2^{-10}, 2^{-9.5}, 2^{-9}\}$ \\
$1\text{e}{15}$ & $\{4, 6, 8, 12\}$ & $\{2^{8}, 2^{9}, 2^{10}, 2^{11}, 2^{12}, 2^{13}, 2^{14}\}$ & $\{2^{-12}, 2^{-11.5}, 2^{-11}, 2^{-10.5}, 2^{-10}, 2^{-9.5}, 2^{-9}\}$ \\
$3\text{e}{15}$ & $\{6, 8, 12, 16\}$ & $\{2^{8}, 2^{9}, 2^{10}, 2^{11}, 2^{12}, 2^{13}, 2^{14}\}$ & $\{2^{-12}, 2^{-11.5}, 2^{-11}, 2^{-10.5}, 2^{-10}, 2^{-9.5}, 2^{-9}\}$ \\
$1\text{e}{16}$ & $\{6, 8, 12, 16\}$ & $\{2^{9}, 2^{10}, 2^{11}, 2^{12}, 2^{13}, 2^{14}, 2^{15}\}$ & $\{2^{-12}, 2^{-11.5}, 2^{-11}, 2^{-10.5}, 2^{-10}, 2^{-9.5}, 2^{-9}\}$ \\
$3\text{e}{16}$ & $\{8, 12, 16, 24\}$ & $\{2^{9}, 2^{10}, 2^{11}, 2^{12}, 2^{13}, 2^{14}, 2^{15}\}$ & $\{2^{-12}, 2^{-11.5}, 2^{-11}, 2^{-10.5}, 2^{-10}, 2^{-9.5}, 2^{-9}\}$ \\
$1\text{e}{17}$ & $\{12, 16, 24, 32\}$ & $\{2^{10}, 2^{11}, 2^{12}, 2^{13}, 2^{14}, 2^{15}, 2^{16}\}$ & $\{2^{-12}, 2^{-11.5}, 2^{-11}, 2^{-10.5}, 2^{-10}, 2^{-9.5}, 2^{-9}\}$ \\
\bottomrule
\end{tabular}
\caption{Grid search configurations for fitting hyperparameter scaling laws in Craftax.}
\label{tab:craftax_hparam_sweep}
\end{table}

\begin{table}[ht]
\centering
\small
\begin{tabular}{llll}
\toprule
FLOPs & Model widths & Batch size & Learning rate \\
\midrule
$1\text{e}{15}$ & $\{4, 5, 6, 7\}$ & $2^{9}$ & $2^{-9.12}$ \\
$3\text{e}{15}$ & $\{5, 6, 7, 8, 9, 10\}$ & $2^{10}$ & $2^{-8.97}$ \\
$1\text{e}{16}$ & $\{6, 7, 8, 9, 10, 11, 12, 14\}$ & $2^{10}$ & $2^{-9.35}$ \\
$3\text{e}{16}$ & $\{8, 10, 12, 14, 16, 18, 20\}$ & $2^{11}$ & $2^{-9.19}$ \\
$1\text{e}{17}$ & $\{10, 12, 14, 16, 18, 20, 24\}$ & $2^{12}$ & $2^{-9.07}$ \\
$1\text{e}{18}$ & $\{16, 20, 24, 26, 28, 34, 40\}$ & $2^{13}$ & $2^{-9.30}$ \\
$1\text{e}{19}$ & $\{43\}$ & $2^{14}$ & $2^{-9.53}$ \\
\bottomrule
\end{tabular}
\caption{Configurations for the isoFLOPs profile experiment in Starpilot.}
\label{tab:isoflops_procgen}
\end{table}

\begin{table}[ht]
\centering
\small
\begin{tabular}{llll}
\toprule
FLOPs & Model widths & Batch size & Learning rate \\
\midrule
$1\text{e}{15}$ & $\{5, 6, 7, 8, 9, 10, 11, 12\}$ & $2^{11}$ & $2^{-10.17}$ \\
$3\text{e}{15}$ & $\{6, 7, 8, 9, 10, 11, 12, 14, 16\}$ & $2^{11}$ & $2^{-10.51}$ \\
$1\text{e}{16}$ & $\{7, 8, 9, 10, 11, 12, 13, 14, 16, 18, 20\}$ & $2^{12}$ & $2^{-10.38}$ \\
$3\text{e}{16}$ & $\{10, 12, 14, 16, 18, 20, 24, 28\}$ & $2^{12}$ & $2^{-10.71}$ \\
$1\text{e}{17}$ & $\{16, 18, 20, 22, 24, 26, 28, 30, 32, 34\}$ & $2^{13}$ & $2^{-10.58}$ \\
$1\text{e}{18}$ & $\{24, 28, 30, 32, 34, 35, 36, 40, 44, 48\}$ & $2^{14}$ & $2^{-10.79}$ \\
$1\text{e}{19}$ & $\{58\}$ & $2^{15}$ & $2^{-11.00}$ \\
\bottomrule
\end{tabular}
\caption{Configurations of the isoFLOPs profile experiment in Craftax.}
\label{tab:isoflops_craftax}
\end{table}

%% file: iclr2027_conference.bib
@article{ba2016layer,
  title={Layer normalization},
  author={Ba, Jimmy Lei and Kiros, Jamie Ryan and Hinton, Geoffrey E},
  journal={arXiv preprint arXiv:1607.06450},
  year={2016}
}

@article{berner2019dota,
  title={Dota 2 with large scale deep reinforcement learning},
  author={Berner, Christopher and Brockman, Greg and Chan, Brooke and Cheung, Vicki and Dennison, Christy and Farhi, David and Fischer, Quirin and Hashme, Shariq and Hesse, Chris and J{\u{A}}{\l}zefowicz, Rafal and others},
  journal={arXiv preprint arXiv:1912.06680},
  year={2019}
}

@article{bi2024deepseek,
  title={Deepseek llm: Scaling open-source language models with longtermism},
  author={Bi, Xiao and Chen, Deli and Chen, Guanting and Chen, Shanhuang and Dai, Damai and Deng, Chengqi and Ding, Honghui and Dong, Kai and Du, Qiushi and Fu, Zhe and others},
  journal={arXiv preprint arXiv:2401.02954},
  year={2024}
}

@article{Bradley1952RankAO,
  title={Rank Analysis of Incomplete Block Designs: I. The Method of Paired Comparisons},
  author={Ralph Allan Bradley and Milton E. Terry},
  journal={Biometrika},
  volume={39},
  number={3/4},
  pages={324--345},
  year={1952},
  publisher={[Oxford University Press, Biometrika Trust]}
}

@article{cheng2026isocompute,
  title={IsoCompute Playbook: Optimally Scaling Sampling Compute for LLM RL},
  author={Cheng, Zhoujun and Xie, Yutao and Qu, Yuxiao and Setlur, Amrith and Hao, Shibo and Pimpalkhute, Varad and Liang, Tongtong and Yao, Feng and Liu, Zhengzhong and Xing, Eric and others},
  journal={arXiv preprint arXiv:2603.12151},
  year={2026}
}

@inproceedings{cobbe2020leveraging,
  title={Leveraging procedural generation to benchmark reinforcement learning},
  author={Cobbe, Karl and Hesse, Chris and Hilton, Jacob and Schulman, John},
  booktitle={International conference on machine learning},
  pages={2048--2056},
  year={2020},
  organization={PMLR}
}

@article{comanici2025gemini,
  title={Gemini 2.5: Pushing the frontier with advanced reasoning, multimodality, long context, and next generation agentic capabilities},
  author={Comanici, Gheorghe and Bieber, Eric and Schaekermann, Mike and Pasupat, Ice and Sachdeva, Noveen and Dhillon, Inderjit and Blistein, Marcel and Ram, Ori and Zhang, Dan and Rosen, Evan and others},
  journal={arXiv preprint arXiv:2507.06261},
  year={2025}
}

@book{elo1978rating,
  address = {New York},
  author = {Elo, Arpad E.},
  publisher = {Arco Pub.},
  title = {The Rating of Chessplayers, Past and Present},
  year = {1978},
  isbn = {0668047216}
}

@inproceedings{espeholt2018impala,
  title={Impala: Scalable distributed deep-rl with importance weighted actor-learner architectures},
  author={Espeholt, Lasse and Soyer, Hubert and Munos, Remi and Simonyan, Karen and Mnih, Vlad and Ward, Tom and Doron, Yotam and Firoiu, Vlad and Harley, Tim and Dunning, Iain and others},
  booktitle={International conference on machine learning},
  pages={1407--1416},
  year={2018},
  organization={PMLR}
}

@article{fu2026compute,
  title={Compute-optimal scaling for value-based deep rl},
  author={Fu, Preston and Rybkin, Oleh and Zhou, Zhiyuan Paul and Nauman, Michal and Abbeel, Pieter and Levine, Sergey and Kumar, Aviral},
  journal={Advances in Neural Information Processing Systems},
  volume={38},
  pages={160363--160402},
  year={2026}
}

@article{guo2025deepseek,
  title={Deepseek-r1: Incentivizing reasoning capability in llms via reinforcement learning},
  author={Guo, Daya and Yang, Dejian and Zhang, Haowei and Song, Junxiao and Wang, Peiyi and Zhu, Qihao and Xu, Runxin and Zhang, Ruoyu and Ma, Shirong and Bi, Xiao and others},
  journal={arXiv preprint arXiv:2501.12948},
  year={2025}
}

@softwareversion{hamon:hal-04659863v1,
  TITLE = {{transformerXL\_PPO\_JAX}},
  AUTHOR = {Hamon, Gautier},
  URL = {https://inria.hal.science/hal-04659863},
  NOTE = {},
  YEAR = {2024},
  MONTH = Jul,
  REPOSITORY = {https://github.com/Reytuag/transformerXL_PPO_JAX},
  LICENSE = {MIT License},
  HAL_ID = {hal-04659863},
  HAL_VERSION = {v1},
}

@inproceedings{he2016deep,
  title={Deep residual learning for image recognition},
  author={He, Kaiming and Zhang, Xiangyu and Ren, Shaoqing and Sun, Jian},
  booktitle={Proceedings of the IEEE conference on computer vision and pattern recognition},
  pages={770--778},
  year={2016}
}

@inproceedings{henderson2018deep,
  title={Deep reinforcement learning that matters},
  author={Henderson, Peter and Islam, Riashat and Bachman, Philip and Pineau, Joelle and Precup, Doina and Meger, David},
  booktitle={Proceedings of the AAAI conference on artificial intelligence},
  volume={32},
  year={2018}
}

@article{Hendrycks2016GaussianEL,
  title={Gaussian Error Linear Units (GELUs)},
  author={Dan Hendrycks and Kevin Gimpel},
  journal={arXiv: Learning},
  year={2016},
  url={https://arxiv.org/abs/1606.08415}
}

@article{henighan2020scaling,
  title={Scaling laws for autoregressive generative modeling},
  author={Henighan, Tom and Kaplan, Jared and Katz, Mor and Chen, Mark and Hesse, Christopher and Jackson, Jacob and Jun, Heewoo and Brown, Tom B and Dhariwal, Prafulla and Gray, Scott and others},
  journal={arXiv preprint arXiv:2010.14701},
  year={2020}
}

@article{hestness2017deep,
  title={Deep learning scaling is predictable, empirically},
  author={Hestness, Joel and Narang, Sharan and Ardalani, Newsha and Diamos, Gregory and Jun, Heewoo and Kianinejad, Hassan and Patwary, Md Mostofa Ali and Yang, Yang and Zhou, Yanqi},
  journal={arXiv preprint arXiv:1712.00409},
  year={2017}
}

@article{hilton2023scaling,
  title={Scaling laws for single-agent reinforcement learning},
  author={Hilton, Jacob and Tang, Jie and Schulman, John},
  journal={arXiv preprint arXiv:2301.13442},
  year={2023}
}

@article{hochreiter1997long,
  title={Long short-term memory},
  author={Hochreiter, Sepp and Schmidhuber, J{\"u}rgen},
  journal={Neural computation},
  volume={9},
  number={8},
  pages={1735--1780},
  year={1997},
  publisher={MIT press}
}

@article{hoffmann2022training,
  title={Training compute-optimal large language models},
  author={Hoffmann, Jordan and Borgeaud, Sebastian and Mensch, Arthur and Buchatskaya, Elena and Cai, Trevor and Rutherford, Eliza and Casas, Diego de Las and Hendricks, Lisa Anne and Welbl, Johannes and Clark, Aidan and others},
  journal={arXiv preprint arXiv:2203.15556},
  year={2022}
}

@article{jaech2024openai,
  title={Openai o1 system card},
  author={Jaech, Aaron and Kalai, Adam and Lerer, Adam and Richardson, Adam and El-Kishky, Ahmed and Low, Aiden and Helyar, Alec and Madry, Aleksander and Beutel, Alex and Carney, Alex and others},
  journal={arXiv preprint arXiv:2412.16720},
  year={2024}
}

@article{jones2021scaling,
  title={Scaling scaling laws with board games},
  author={Jones, Andy L},
  journal={arXiv preprint arXiv:2104.03113},
  year={2021}
}

@article{kaplan2020scaling,
  title={Scaling laws for neural language models},
  author={Kaplan, Jared and McCandlish, Sam and Henighan, Tom and Brown, Tom B and Chess, Benjamin and Child, Rewon and Gray, Scott and Radford, Alec and Wu, Jeffrey and Amodei, Dario},
  journal={arXiv preprint arXiv:2001.08361},
  year={2020}
}

@inproceedings{khatri2026art,
  title={The art of scaling reinforcement learning compute for llms},
  author={Khatri, Devvrit and Madaan, Lovish and Tiwari, Rishabh and Bansal, Rachit and Duvvuri, Venkata Sai Surya Subramanyam and Zaheer, Manzil and Dhillon, Inderjit and Brandfonbrener, David and Agarwal, Rishabh},
  booktitle={International Conference on Learning Representations},
  volume={2026},
  pages={72438--72467},
  year={2026}
}

@inproceedings{kingma2015adam,
  author = {Kingma, Diederik P. and Ba, Jimmy},
  booktitle = {ICLR (Poster)},
  editor = {Bengio, Yoshua and LeCun, Yann},
  ee = {http://arxiv.org/abs/1412.6980},
  title = {Adam: A Method for Stochastic Optimization.},
  url = {http://dblp.uni-trier.de/db/conf/iclr/iclr2015.html#KingmaB14},
  year = 2015
}

@article{malladi2022sdes,
  title={On the SDEs and scaling rules for adaptive gradient algorithms},
  author={Malladi, Sadhika and Lyu, Kaifeng and Panigrahi, Abhishek and Arora, Sanjeev},
  journal={Advances in Neural Information Processing Systems},
  volume={35},
  pages={7697--7711},
  year={2022}
}

@article{muennighoff2023scaling,
  title={Scaling data-constrained language models},
  author={Muennighoff, Niklas and Rush, Alexander and Barak, Boaz and Le Scao, Teven and Tazi, Nouamane and Piktus, Aleksandra and Pyysalo, Sampo and Wolf, Thomas and Raffel, Colin A},
  journal={Advances in Neural Information Processing Systems},
  volume={36},
  pages={50358--50376},
  year={2023}
}

@inproceedings{neumann2023scaling,
title={Scaling Laws for a Multi-Agent Reinforcement Learning Model},
author={Oren Neumann and Claudius Gros},
booktitle={The Eleventh International Conference on Learning Representations },
year={2023},
url={https://openreview.net/forum?id=ZrEbzL9eQ3W}
}

@InProceedings{pmlr-v235-matthews24a,
  title = 	 {Craftax: A Lightning-Fast Benchmark for Open-Ended Reinforcement Learning},
  author =       {Matthews, Michael and Beukman, Michael and Ellis, Benjamin and Samvelyan, Mikayel and Jackson, Matthew Thomas and Coward, Samuel and Foerster, Jakob Nicolaus},
  booktitle = 	 {Proceedings of the 41st International Conference on Machine Learning},
  pages = 	 {35104--35137},
  year = 	 {2024},
  editor = 	 {Salakhutdinov, Ruslan and Kolter, Zico and Heller, Katherine and Weller, Adrian and Oliver, Nuria and Scarlett, Jonathan and Berkenkamp, Felix},
  volume = 	 {235},
  series = 	 {Proceedings of Machine Learning Research},
  month = 	 {21--27 Jul},
  publisher =    {PMLR},
  url = 	 {https://proceedings.mlr.press/v235/matthews24a.html}
}

@InProceedings{pmlr-v48-mniha16,
  title = 	 {Asynchronous Methods for Deep Reinforcement Learning},
  author = 	 {Mnih, Volodymyr and Badia, Adria Puigdomenech and Mirza, Mehdi and Graves, Alex and Lillicrap, Timothy and Harley, Tim and Silver, David and Kavukcuoglu, Koray},
  booktitle = 	 {Proceedings of The 33rd International Conference on Machine Learning},
  pages = 	 {1928--1937},
  year = 	 {2016},
  editor = 	 {Balcan, Maria Florina and Weinberger, Kilian Q.},
  volume = 	 {48},
  series = 	 {Proceedings of Machine Learning Research},
  address = 	 {New York, New York, USA},
  month = 	 {20--22 Jun},
  publisher =    {PMLR},
  url = 	 {https://proceedings.mlr.press/v48/mniha16.html}
}

@inproceedings{rosenfeld2020a,
title={A Constructive Prediction of the Generalization Error Across Scales},
author={Jonathan S. Rosenfeld and Amir Rosenfeld and Yonatan Belinkov and Nir Shavit},
booktitle={International Conference on Learning Representations},
year={2020},
url={https://openreview.net/forum?id=ryenvpEKDr}
}

@article{schaul2019ray,
  title={Ray interference: a source of plateaus in deep reinforcement learning},
  author={Schaul, Tom and Borsa, Diana and Modayil, Joseph and Pascanu, Razvan},
  journal={arXiv preprint arXiv:1904.11455},
  year={2019}
}

@article{silver2017mastering,
  title={Mastering the game of go without human knowledge},
  author={Silver, David and Schrittwieser, Julian and Simonyan, Karen and Antonoglou, Ioannis and Huang, Aja and Guez, Arthur and Hubert, Thomas and Baker, Lucas and Lai, Matthew and Bolton, Adrian and others},
  journal={nature},
  volume={550},
  number={7676},
  pages={354--359},
  year={2017},
  publisher={Nature Publishing Group UK London}
}

@inproceedings{tan2026scaling,
  title={Scaling behaviors of llm reinforcement learning post-training: An empirical study in mathematical reasoning},
  author={Tan, Zelin and Geng, Hejia and Yu, Xiaohang and Zhang, Mulei and Wan, Guancheng and Zhou, Yifan and He, Qiang and Xue, Xiangyuan and Zhou, Heng and Fan, Yutao and others},
  booktitle={Proceedings of the 64th Annual Meeting of the Association for Computational Linguistics (Volume 1: Long Papers)},
  pages={31300--31319},
  year={2026}
}

@article{vinyals2019grandmaster,
  title={Grandmaster level in StarCraft II using multi-agent reinforcement learning},
  author={Vinyals, Oriol and Babuschkin, Igor and Czarnecki, Wojciech M and Mathieu, Micha{\"e}l and Dudzik, Andrew and Chung, Junyoung and Choi, David H and Powell, Richard and Ewalds, Timo and Georgiev, Petko and others},
  journal={nature},
  volume={575},
  number={7782},
  pages={350--354},
  year={2019},
  publisher={Nature Publishing Group UK London}
}
